\documentclass[lettersize,journal]{IEEEtran}
\usepackage{amsmath,amsfonts}
\usepackage{algorithmic}
\usepackage{algorithm}
\usepackage{array}
\usepackage[caption=false,font=normalsize,labelfont=sf,textfont=sf]{subfig}
\usepackage{textcomp}
\usepackage{stfloats}
\usepackage{url}
\usepackage{graphicx}
\usepackage{cite}
\usepackage{csquotes}
\usepackage[english]{babel}
\usepackage{tikz}
\usetikzlibrary{positioning}
\usepackage{booktabs}
\usepackage{siunitx}
\usepackage{makecell}
\usepackage{multirow}
\usepackage[final]{changes}
\usepackage{threeparttable}
\usepackage{hyperref}
\usepackage[percent]{overpic} 
\usepackage{xcolor}

\usepackage{soul}
\newcommand{\minoradd}[1]{\textcolor{blue}{#1}}
\newcommand{\minordel}[1]{\textcolor{blue}{\st{#1}}}

\renewcommand{\minoradd}[1]{#1}
\renewcommand{\minordel}[1]{}

\newcommand\copyrighttext{%
	\footnotesize \copyright 2025 IEEE. Personal use of this material is permitted. Permission from IEEE must be obtained for all other uses, in any current or future media, including reprinting/republishing this material for advertising or promotional purposes, creating new collective works, for resale or redistribution to servers or lists, or reuse of any copyrighted component of this work in other works.}
\newcommand\copyrightnotice{%
	\begin{tikzpicture}[remember picture,overlay]
		\node[anchor=south,yshift=5pt] at (current page.south) {\fbox{\parbox{\dimexpr\textwidth-\fboxsep-\fboxrule\relax}{\copyrighttext}}};
	\end{tikzpicture}%
}

\begin{document}

\title{A Communication-Latency-Aware Co-Simulation Platform for Safety and Comfort Evaluation of Cloud-Controlled ICVs}

\author{Yongqi Zhao, Xinrui Zhang, Tomislav Mihalj, Martin Schabauer, Luis Putzer, Erik Reichmann-Blaga, Ádám Boronyák, András Rövid, Gábor Soós, Peizhi Zhang, Lu Xiong, Jia Hu~\IEEEmembership{Senior Member,~IEEE}, and Arno Eichberger~\IEEEmembership{Member,~IEEE}
        
\thanks{This work was supported by the National Key R\&D Program of China under Grant Nr. 2022YFE0117100, by the FFG in the research project PECOP (FFG Projektnummer 893988), as part of the~\enquote{Bilateral Cooperation Austria - People’s Republic of China / MOST 2nd Call} program, and by FFG in the research project Central System (FFG Project number 39067121). (\textit{Corresponding author: Xinrui Zhang})}
\thanks{Yongqi Zhao is with the School of Automotive Studies, Tongji University, Shanghai 201804, China and the Institute of Automotive Engineering, Graz University of Technology, Graz 8010, Austria (e-mail: yongqi.zhao@tugraz.at)}
\thanks{Xinrui Zhang, Peizhi Zhang, and Lu Xiong are with the School of Automotive Studies, Tongji University, Shanghai 201804, China (e-mail: zhangxr@tongji.edu.cn; zhangpeizhitom@126.com; xiong$\_$lu@tongji.edu.cn).}
\thanks{Tomislav Mihalj, Martin Schabauer, Luis Putzer, Erik Reichmann-Blaga, and Arno Eichberger are with the Institute of Automotive Engineering, Graz University of Technology, Graz 8010, Austria (e-mail: tomislav.mihalj@tugraz.at; martin.schabauer@tugraz.at; putzer@alumni.tugraz.at; e.reichmannblaga@gmail.com; arno.eichberger@tugraz.at).}
\thanks{Ádám Boronyák and András Rövid are with Department of Automotive Technologies, Budapest University of Technology and Economics, Budapest, 1111, Hungary (e-mail: boronyak.adam@kjk.bme.hu; rovid.andras@kjk.bme.hu).}
\thanks{Gábor Soós is with Magyar Telekom Nyrt., Budapest, 1097, Hungary (e-mail: soos.gabor2@telekom.hu).}
\thanks{Jia Hu is with Key Laboratory of Road and Traffic Engineering of the
Ministry of Education, Tongji University, Shanghai 201804, China (e-mail:
hujia@tongji.edu.cn).}


}

\markboth{Journal of \LaTeX\ Class Files,~Vol.~14, No.~8, August~2021}%
{Shell \MakeLowercase{\textit{et al.}}: A Sample Article Using IEEEtran.cls for IEEE Journals}



\maketitle

\begin{abstract}
Testing cloud-controlled intelligent connected vehicles (ICVs) requires simulation environments that faithfully emulate both vehicle behavior and realistic communication latencies. This paper proposes a latency-aware co-simulation platform integrating CarMaker and Vissim to evaluate safety and comfort under real-world vehicle-to-cloud (V2C) latency conditions. \replaced{Three}{Two} communication latency models, derived from empirical 5G measurements in China and Hungary, are incorporated and statistically modeled using Gamma distributions. A proactive conflict module (PCM) is proposed to dynamically control background vehicles and generate safety-critical scenarios. The platform is validated through experiments involving an exemplary system under test (SUT) across \replaced{eight}{six} testing conditions combining two PCM modes (enabled/disabled) and \replaced{four}{three} latency conditions (none, China, Hungary\added{, abnormal}). Safety and comfort are assessed using metrics including collision rate, distance headway, post-encroachment time, and the spectral characteristics of longitudinal acceleration. Results show that the PCM effectively increases driving environment criticality, \replaced{while V2C latency reduces ride comfort and, under extreme driving conditions, further aggravates safety-critical scenarios.}{while V2C latency primarily affects ride comfort.} These findings confirm the platform's effectiveness in systematically evaluating cloud-controlled ICVs under diverse testing conditions. 
\end{abstract}


\section{Introduction}\copyrightnotice
Intelligent connected vehicles (ICVs) utilize vehicle-to-everything (V2X) technology to enable cooperative driving, which enhances traffic efficiency and improves the overall driving experience~\cite{SAE_2021}. Among these, cloud-controlled ICVs further smooth traffic flow through integration of real-time data exchange with cloud platforms~\cite{chu2021cloud}. To support the deployment of such technologies, various countries have introduced strategic initiatives, such as the U.S. Department of Transportation’s~\enquote{Saving Lives with Connectivity} program~\cite{USDOT_2024}. Nevertheless, the large-scale deployment of ICVs requires rigorous testing to ensure their safety and reliability.

At present, three main methods are employed to test ICVs, namely on-road, closed-track, and simulation testing~\cite{11164300}. Although on-road testing provides realism, it is inefficient due to the randomness and uncontrollability of real-world scenarios, requiring billions of miles to sufficiently validate ICVs safety and reliability~\cite{KALRA2016182}. Closed-track testing, while efficient and realistic, requires considerable investment in specialized facilities; for example, the ZalaZONE facility in Hungary was built at a cost of 159 million dollars~\cite{arminas2020zalazone}. Simulation testing, which combines efficiency, cost-effectiveness, and realism (when close to reality), has become essential in ICVs development, as exemplified by its significant role in advancing Waymo’s technology~\cite{cerf2018cartest}. However, rapid advancements in V2X communication for ICVs present evolving challenges for simulation testing. 

The testing scope for ICVs has evolved from focusing on individual vehicles to encompassing vehicle-network-cloud systems, where control is managed via cloud platforms. Cloud control involves managing motion planning and control in the cloud, utilizing environmental perception and positioning data from roadside infrastructure and ICVs, as well as transmitting control signals through the network~\cite{chu2021cloud}. This architecture requires simulation testing that incorporates vehicle dynamics, driving environments, and network communication latency.

In recent decades, the development of simulators for automated driving has primarily focused on single specific aspects such as traffic flow, sensory data, driving policy or vehicle dynamics, often lacking V2X capabilities~\cite{10461065}. For example, Vissim models realistic traffic flows, but lacks vehicle dynamics, while vehicle dynamics simulators like CarMaker offer detailed vehicle dynamics but cannot simulate traffic flows. To address these limitations and incorporate V2X communication, various software combinations have been proposed, such as NS-3 with Vissim~\cite{7795680, CHOUDHURY20162042} or CarMaker~\cite{9569386}, and OMNeT\texttt{++}\footnote{\url{https://omnetpp.org/}} with SUMO~\cite{JIA2021102984} or CARLA~\cite{10136319}. However, these solutions focus on cooperative control systems, emphasizing the improvements of V2X on the traffic efficiency. Due to the scarcity of real-world measurement data, communication latency is frequently treated in a simplified manner.

In addition to software, hardware components have also been integrated into simulation platforms to enhance fidelity. Real test vehicles~\cite{eichberger2017car2x, FENG2020105664} and driving simulators~\cite{electronics9101645,dong2025human} simulate the vehicle dynamics, while augmented reality, traffic simulator, and real traffic are used to model background vehicles~\cite{FENG2020105664, electronics9101645, eichberger2017car2x, 9576134}. Cohda wireless device and V2X signal generator have been utilized to establish V2X communication~\cite{eichberger2017car2x, electronics9101645, 8814263, s22135019}.

\begin{table*}[ht]
\caption{\added{Comparison of previous simulation platforms and the proposed latency-aware co-simulation platform.}}
\label{tab:work_comparison}
\centering
\begin{tabular}{lcccc}\toprule[2pt]
\added{Work} & \added{Vehicle dynamics} & \added{Background vehicle (BGV)} & \added{BGV control} & \added{Communication latency} \\
\midrule[1pt]
\added{\cite{7795680}}  & \added{Point mass model} & \added{-} & \added{-} & \added{Network simulator-based model}\\
\added{\cite{CHOUDHURY20162042}}  & \added{Point mass model} & \added{Synthetic traffic} & \added{-} & \added{Network simulator-based model}\\
\added{\cite{9569386}}  & \added{Sample multi-body model} & \added{Synthetic traffic} & \added{-} & \added{Network simulator-based model}\\
\added{\cite{JIA2021102984}}  & \added{Sample bicycle model} & \added{Synthetic traffic} & \added{$\surd$} & \added{Network simulator-based model}\\
\added{\cite{10136319}}  & \added{Sample bicycle model} & \added{Synthetic traffic} & \added{-} & \added{Network simulator-based model}\\
\added{\cite{eichberger2017car2x}}  & \added{Sample multi-body model} & \added{Synthetic traffic} & \added{-} & \added{Fixed-value model}\\
\added{\cite{FENG2020105664}}  & \added{Real vehicle} & \added{Synthetic traffic} & \added{-} & \added{-}\\
\added{\cite{electronics9101645}}  & \added{Sample multi-body model} & 
\added{Synthetic traffic} & - & \added{Hardware-in-the-loop model}\\
\cite{dong2025human} & \minoradd{Sample multi-body model} & \minoradd{Synthetic traffic} & \minoradd{-} & \minoradd{-}\\
\added{\cite{s22135019}}  & \added{Sample multi-body model} & \added{Synthetic traffic} & \added{-} & \added{Hardware-in-the-loop model}\\
\added{\textbf{Ours}}  & \added{\textbf{Measurement-calibrated multi-body model}} & \added{\textbf{Measurement-calibrated traffic}} & \added{$\surd$} & \added{\textbf{Measurement-based probabilistic model}}\\
\bottomrule[2pt]
\end{tabular}
\vspace{2pt}
\noindent{\footnotesize{*\added{$\surd$ indicates support for active BGV control. \enquote{Fixed-value model} indicates a constant latency assumption.}}}
\end{table*}

In summary, three limitations are evident in existing simulation platforms. First, vehicle dynamics and traffic flow are often handled separately, with few platforms integrating both and frequently lacking V2X simulation, which makes ICVs testing infeasible. Second, the absence of real-world communication latency data has constrained research on its effects. Third, testing efficiency is limited, as the ego vehicle is generally exposed to a low critical driving environment, reducing interaction with challenging scenarios.

To address the identified limitations, a co-simulation platform has been developed with the following contributions:
\begin{enumerate}
    \item Simulations of vehicle dynamics, traffic flow, and vehicle-to-cloud (V2C) communication are achieved by integrating CarMaker and Vissim.
    \item The vehicle dynamics model and traffic flow are built upon real measurements~\added{and integrated into a co-simulation platform to enhance realism.}~\deleted{, while two}\added{Two} V2C communication latency models, based on data collected in China and Hungary, are implemented to represent realistic communication latencies for different use cases. 
    \item The V2C communication latency distributions are statistically fitted to measurement data and further validated through analytical derivation to follow Gamma distributions.
    \item Background vehicles (BGVs) are strategically controlled to ensure higher exposure to critical driving scenarios, thereby enhancing testing efficiency. 
    \item The effects of varying V2C communication latencies on ICVs are systematically investigated and analyzed.
\end{enumerate}

A tool is described for simulating critical driving scenarios with realistic communication latencies, aiding the development of robust ICVs algorithms. Safety and comfort performance under diverse latency conditions is also facilitated.~\added{Table~\ref{tab:work_comparison} provides a comparison between the proposed platform and representative prior works across key aspects.}

\section{Related Work}
\added{According to~\cite{10461065},}~\deleted{S}\added{s}ince the 1990s, simulation platforms tailored for the development and validation of automated driving systems (ADSs) have gained increasing importance. They offer significant advantages by reducing the need for extensive on-road testing and accelerating the verification and validation process. Over the past three decades, numerous simulators have been developed for various purposes, including traffic control design, sensor data processing, driving policy formulation, vehicle dynamics optimization, and vehicle control. However, these simulators are typically specialized, often focusing on a single aspect, such as vehicle dynamics or traffic flow, without providing integrated solutions. Furthermore, contributions to the simulation of V2X communication have been relatively limited.\deleted{\mbox{\cite{10461065}}}

In response to these limitations, co-simulation frameworks have been proposed, which integrate multiple tools for more comprehensive simulations. Most frameworks combine a mainstream simulator (e.g., Vissim, SUMO, CARLA, or CarMaker) with a third-party software to address specific gaps. For example, Vissim, proficient in simulating traffic flow but lacking V2X functionality, was integrated with MATLAB for platoon control and NS-3 for communication constraints~\cite{7795680, CHOUDHURY20162042}. Similarly, SUMO was combined with TraaS~\cite{s18124399}, and OMNeT\texttt{++}~\cite{JIA2021102984} to enable V2X simulation. CARLA was extended through Python API expansion~\cite{8892860, 10328084}, OMNeT\texttt{++}~\cite{10136319}, and Artery V2X~\cite{10575990} for V2X functionality. Other notable efforts include OpenCDA, which integrates CARLA with SUMO~\cite{9564825}, and a framework combining NS-3 and CarMaker for V2X simulation~\cite{9569386}. Nalic et al.~\cite{8916839} developed a co-simulation framework to generate critical scenarios, yet not supporting V2X functionality. Additionally, Gazebo was employed to build V2X-compatible simulation platforms~\cite{9294660}, and~\cite{10508262} recently introduced a simulation platform for truck platoon management.

In addition to software-only simulations, vehicle-in-the-loop (ViL) approach was integrated to enhance simulation realism. Eichberger et al.~\cite{eichberger2017car2x} employed a vehicle equipped with a Cohda MK4 device to gather V2X communication data and develop a sensor model, while Peters et al.~\cite{9576134} used a Cohda MK6\footnote{\url{https://www.cohdawireless.com/solutions/mk6/}} device to establish a C-V2X platform. Lei et al.~\cite{8814263} designed an in-chamber test scheme using a real vehicle and an CMW 500\footnote{\url{https://www.rohde-schwarz.com/nl/products/test-and-measurement/wireless-tester-network-emulator/rs-cmw500-wideband-radio-communication-tester_63493-10844.html}} signal generator for V2X evaluations. Feng et al.~\cite{FENG2020105664} combined a real vehicle with augmented reality in a ViL setup, although without V2X functionality.

Hardware-in-the-loop (HiL) approaches integrate real hardware into simulation frameworks, offering realistic environments for system testing and validation. Lee et al.~\cite{electronics9101645} developed a HiL simulator based on dSPACE and Cohda MK5 to evaluate cooperative eco-driving systems. Mafakheri et al.~\cite{9448667} implemented real-time HiL simulations with SUMO and a roadside unit to examine interactions between test vehicles and simulated objects. Wang et al.~\cite{s22135019} introduced a laboratory framework combining VTD and RF generator for V2X evaluations, while Gemmi et al.~\cite{gemmi2024colossumoevaluatingcooperativedriving} used SUMO with Colosseum network emulator (cf.~\cite{9677430}) to simulate cooperative, connected, and automated mobility scenarios.

In summary, the reviewed simulation platforms are unable to fully support ICV testing due to their limitations in integrating vehicle dynamics, traffic flow, realistic V2C communication models with latency considerations and efficient exposure to critical driving scenarios. To overcome these drawbacks, a simulation platform combining CarMaker, Vissim and V2C communication latency measurements is proposed in the present work.

\section{Methodology}
\label{sec:method}
\begin{figure}[ht]
    \centering
    \includegraphics[width=\columnwidth]{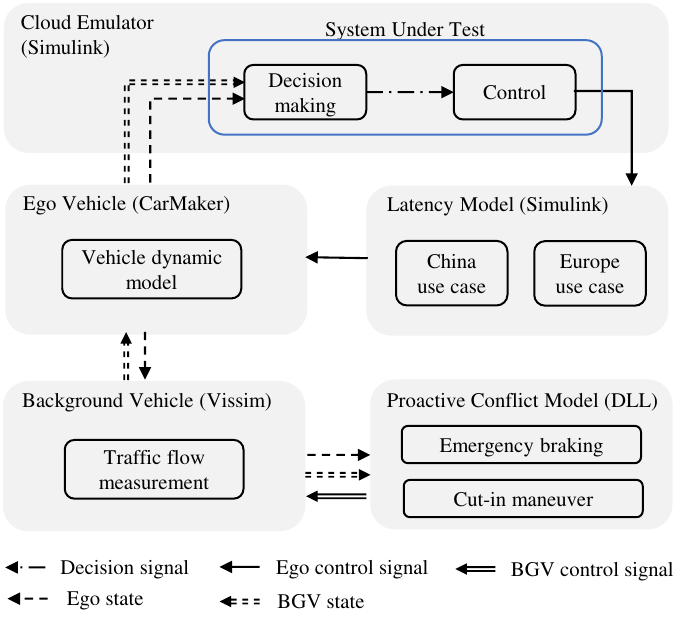}
    \caption{The architecture of the proposed co-simulation platform.}
    \label{fig:ModelArchitecture}
\end{figure}

\autoref{fig:ModelArchitecture} illustrates the architecture of the proposed co-simulation platform. It integrates the system under test (SUT) hosted in a cloud emulator in Simulink environment (cf.~Section~\ref{sec:cloud_emulator}), with a traffic environment modeled in Vissim, where BGVs can be controlled dynamically (cf.~Section~\ref{sec:bgv}). A proactive conflict model (PCM), implemented via a driver model dynamic link library (DLL), enables the manipulation of BGVs in Vissim to generate critical testing scenarios (cf.~Section~\ref{sec:pcm}). The ego vehicle is simulated in CarMaker using a detailed vehicle dynamics model (cf.~Section~\ref{sec:ego_vehicle}). In addition, a V2C communication latency model formulated by integrating empirical fitting from real-world 5G measurements and analytical derivation is included to emulate realistic communication delays (cf.~Section~\ref{sec:latency_model}).

\subsection{Cloud Emulator}
\label{sec:cloud_emulator}
The cloud environment is emulated within Simulink, serving as the deployment platform for the SUT, as illustrated in~\autoref{fig:ModelArchitecture}. The SUT represents a Level 2 ADS, classified according to the SAE taxonomy~\cite{SAE_2018_J3016}. BGVs detection is performed through an ideal sensor installed on the ego vehicle. The term~\enquote{ideal sensor} denotes the acquisition of ground truth data, such as positions and velocities, rather than raw sensor outputs. The states of the detected BGVs and the ego vehicle are transmitted to the cloud in real time. Subsequently, control signals are relayed from the cloud to the ego vehicle, incorporating the effects of communication latency.

\begin{table}[ht]
\caption{Required parameters and characteristics of the vehicle model.}
\label{tab:req_params}
\centering
\begin{tabular}{ll}\toprule[2pt]
Module & Parameters \& characteristics \\
\midrule[1pt]
                  & Mass \\
General  & Moments of inertia\\
                  & Center of gravity\\
\midrule[1pt]
Steering system   & Steering ratio \\
\midrule[1pt]
                  & Spring characteristics\\
Suspension        & Damping characteristics\\
    system        & Anti-roll bar characteristics\\
                  & Suspension kinematics\\
\midrule[1pt]
Tires            & Tire force characteristics \\
\midrule[1pt]
               & Engine characteristics \\
Power train   & Transmission ratio\\
               & Drive system inertia\\
               & Brake system characteristics\\
\bottomrule[2pt]
\end{tabular}
\end{table}

\subsection{Ego Vehicle}
\label{sec:ego_vehicle}
A three-dimensional nonlinear vehicle model of a BMW 640i Gran Coupe (2011), which serves as test vehicle, was built in IPG CarMaker. At first, necessary model parameters and characteristics, see~\autoref{tab:req_params}, were identified by laboratory measurements as well as data sheets and reasonable approximations, in cases where no measurement or data was available. Subsequently, active subsystems, such as active steering, which were not available as template in CarMaker, were modeled separately in MATLAB/Simulink and implemented. Finally, the vehicle model and simulation results were evaluated and experimentally validated by means of measurement data obtained by real-world vehicle dynamics testing conducted on a proving ground. In Appendix~\ref{appendix:vehicle_dynamic_model}, the main topics of building and evaluating the vehicle model are summarized and the validation results are briefly described.

\subsection{Background Vehicle}
\label{sec:bgv}
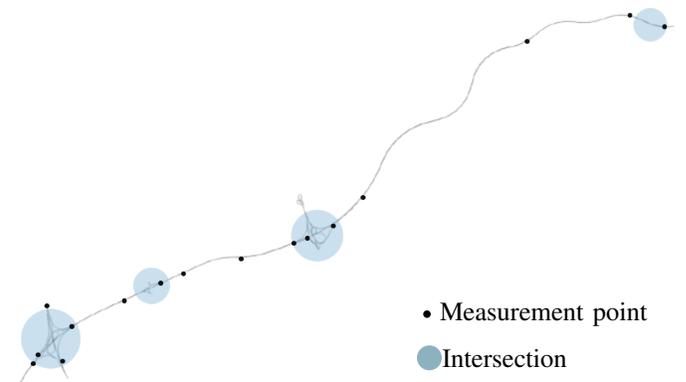
\begin{figure}[ht]
    \centering
    \begin{tikzpicture}
        \node[inner sep=0pt] (image) at (0,0) {\includegraphics[width=\columnwidth]{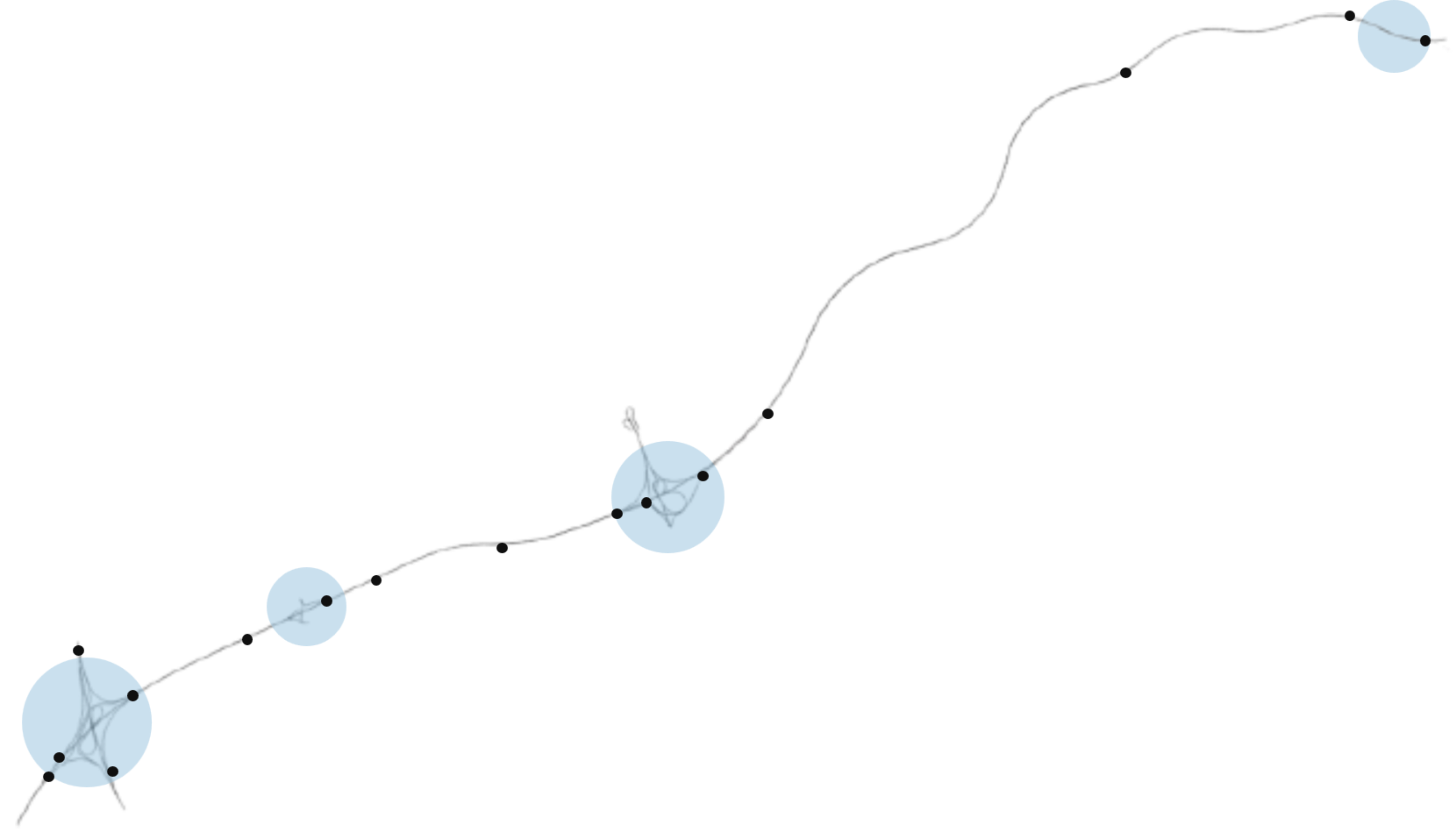}};

        \node[fill=black, opacity=1, circle, minimum size=0.1cm, inner sep=0pt, anchor=north west] (legend1) at (1.05, -1.5) {}; 
        \node[right of=legend1, node distance=1.55cm] {Measurement point};
        
        \node[fill={rgb:red,37;green,150;blue,190}, opacity=0.5, circle, minimum size=0.1cm, anchor=north west] (legend2) at (1, -2) {}; 
        \node[right of=legend2, node distance=1cm] {Intersection};
    \end{tikzpicture}
    \caption{Overview of the road network under test~\cite{8916839}.}
    \label{fig:road_network}
\end{figure}

To accurately represent real traffic density in the proposed simulation platform, comprehensive cross-sectional measurements were conducted across the road network under test depicted in~\autoref{fig:road_network}. Laser scanners and additional measurement devices were placed at various points along the highway, capturing data on vehicle counts, speed distributions, and vehicle classifications. This dataset is essential for formulating the traffic model to reflect naturalistic traffic flow conditions. Further details on the collected data can be found in our previous work~\cite{8916839}.

\added{To translate real-world measurements into Vissim parameters, the laser-scanner and road side sensor data were first processed to extract vehicle trajectories, lane geometries, and speed profiles. These measurements were then aggregated into key traffic flow statistics, including traffic volume, time headway distributions, and vehicle compositions across multiple cross sections of the test road. Based on these statistics, a microscopic traffic demand model in Vissim was calibrated by adjusting vehicle input flows, desired speed distributions, and select Wiedemann'99 car-following parameters to reproduce the observed traffic density and behavior. The calibrated model was further validated by statistical comparison between simulated and measured data. Detailed implementation steps are documented in~\cite{maierhofer2021}.}

\subsection{Proactive Conflict Module}
\label{sec:pcm}
The PCM utilizes the Vissim's driver model application programming interface (API) to create challenging scenarios by customizing BGV behaviors. Through integrating a driver model DLL~in C code, default driving logic is overridden, enabling real-time control of acceleration, braking, and lane changes. Two conflict-inducing behaviors are implemented: an unintended emergency brake by the lead vehicle and a cut-in by an adjacent vehicle. 

\subsubsection{Emergency Braking} A BGV traveling ahead of the ego vehicle in the same lane is designated as the lead vehicle. The distance \( d \) between the two vehicles is computed by
\begin{equation}
d(t) = \left\| \mathbf{p}_{\text{lead}}(t) - \mathbf{p}_{\text{ego}}(t) \right\|,
\label{eq:dst}
\end{equation}
\noindent where \( \mathbf{p}_{\text{lead}}(t) \) and \( \mathbf{p}_{\text{ego}}(t) \) denote the positions of the lead and ego vehicles at time~\(t\), respectively, and $\left\| \cdot \right\|$ denotes the standard Euclidean norm. When its distance to the ego falls below a defined threshold~\( D_{\text{brake}} = 50~m\), i.e., \( d(t) < D_{\text{brake}} \), the lead vehicle applies a sudden deceleration to simulate an emergency brake, thereby assessing the SUT's ability to respond under critical following scenarios.~\minoradd{This threshold is based on the PEGASUS critical highway following distance~\cite{junietz2018pegasus}.}

\subsubsection{Cut-in Maneuver} BGVs in adjacent lanes are evaluated based on their Euclidean distance to the ego vehicle \( d_i \), which is given by
\begin{equation}
d_i(t) = \left\| \mathbf{p}_i(t) - \mathbf{p}_{\text{ego}}(t) \right\|,
\end{equation}
\noindent where \( \mathbf{p}_i(t) \) denotes the position of the \( i \)-th BGV. A BGV is considered a potential cut-in candidate if its distance to the ego vehicle falls below a predefined threshold \( D_{\text{cut}} = 50~m\), i.e., 
\begin{equation}
\mathcal{C} = \{ i \mid d_i(t) < D_{\text{cut}} \}
\end{equation}
\noindent where \( \mathcal{C} \) represents the set of all such candidate BGVs, and \( D_{\text{cut}} \) is the predefined spatial threshold.~\minoradd{This \SI{50}{\meter} value falls between cut-in initiation distance (\SI{30}{\meter}~\cite{lu2024studying}) and the maximum range (\SI{75}{\meter}~\cite{wang2019analysis}).} Among all candidates, the closest BGV is selected by
\begin{equation}
i^* = \arg\min_{i \in \mathcal{C}} d_i(t).
\end{equation}
The selected BGV \( i^* \) then performs a lane-change maneuver into the ego vehicle's lane, introducing a lateral encroachment. This scenario challenges the SUT's perception and decision-making modules in dynamically evolving environments.

\subsection{Latency Model}
\label{sec:latency_model}
Communication latency is often simulated using analytical models, though their reliance on external tools and limited fidelity hinder co-simulation~\cite{9897006, 9964110}. To address this, two probabilistic models are developed using real-world measurements from China and Europe~\cite{ficzere2021compact,ficzere2021a}.

\subsubsection{Overview of 5G Network Architecture}

\begin{figure}[ht]
\centering
\begin{tikzpicture}
    \node[anchor=south west, inner sep=0] (image) at (0,0) {\includegraphics[width=.98\columnwidth]{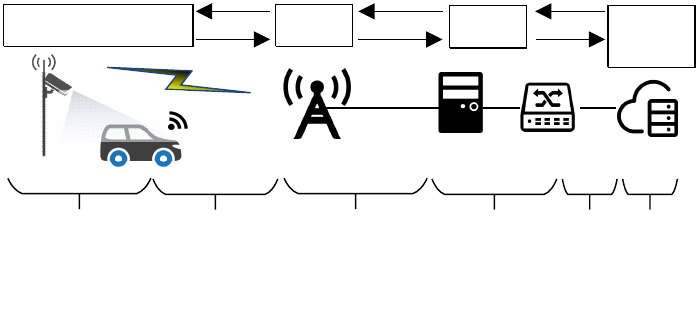}};
    
    \begin{scope}[x={(image.south east)}, y={(image.north west)}]
        \node[] at (0.13, 0.925) {OBU \& RSU};
        \node[] at (0.45, 0.925) {gNB};
        \node[] at (0.7, 0.925) {UPF};
        \node[text width=1cm, align=center, minimum width=1cm, minimum height=0.9cm] at (0.93, 0.889) {App. server};

        \node[] at (0.13, 0.25) {$l_{\text{V2I}}$};
        \node[] at (0.33, 0.25) {$l_{\text{radio}}$};
        \node[] at (0.53, 0.25) {$l_{\text{TN}}$};
        \node[] at (0.73, 0.25) {$l_{\text{CN}}$};
        \node[] at (0.83, 0.25) {$l_{\text{UPF-AS}}$};
        \node[] at (0.93, 0.25) {$l_{\text{AS}}$};

        \draw[dashed] (0.005,0) rectangle (0.22,0.45); 
        \node[text width=2cm, align=center, minimum width=1cm, minimum height=0.9cm, font=\footnotesize] at (0.11, 0.1) {Unique link of Europe case};
        
    \end{scope}
\end{tikzpicture}
\caption{Common network architecture in China and Europe.\label{fig:China_Europe_architecture}}
\end{figure}
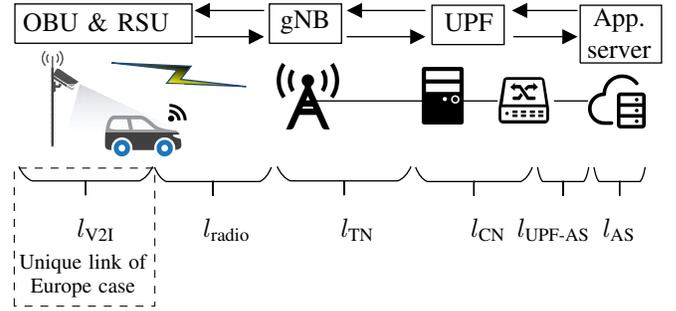

\autoref{fig:China_Europe_architecture} illustrates the 5G network architecture used in the China and Europe cases. In both, the core components include the on-board unit (OBU), next-generation Node B (gNB), user plane function (UPF), and application server (AS), enabling V2C communication. The European architecture further incorporates a vehicle-to-infrastructure (V2I) link via roadside units (RSUs), supporting cooperative perception between vehicle and infrastructure. 

The latency components in~\autoref{fig:China_Europe_architecture} are denoted as~\( l_{\text{V2I}},~l_{\text{radio}},~l_{\text{TN}},~l_{\text{CN}},~l_{\text{UPF-AS}}, \) and~\( l_{\text{AS}} \), representing the delays introduced by V2I communication, wireless transmission, transport network, core network, UPF to AS communication, and AS processing, respectively. The V2I link (\( l_{\text{V2I}} \)) is unique to the Europe use case and enables direct interaction between the ego vehicle and RSUs. Additionally, arrows in the figure represent bidirectional data communication. 

\subsubsection{China Use Case}

\subsubsection*{Overview of V2X Solution} 
\begin{figure}[ht]
    \centering
    \includegraphics[width=0.98\columnwidth]{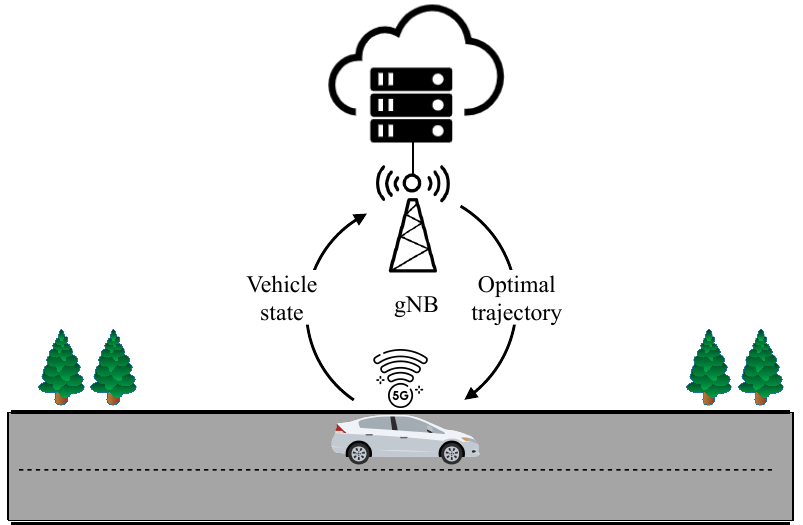}
    \caption{V2X communication solution in China use case.}
    \label{fig:The_data_interaction_process}
\end{figure}

The process of vehicle-cloud interaction is illustrated in~\autoref{fig:The_data_interaction_process}. Real-time vehicle state data are transmitted to a 5G base station (gNB) and routed through the core network to a cloud control center, where optimized trajectories are computed and sent back to the vehicle via the same 5G infrastructure. 

\subsubsection*{Probabilistic Model} 
\begin{figure*}[ht]
  \centering
  \subfloat[]{%
    \includegraphics[width=0.23\linewidth]{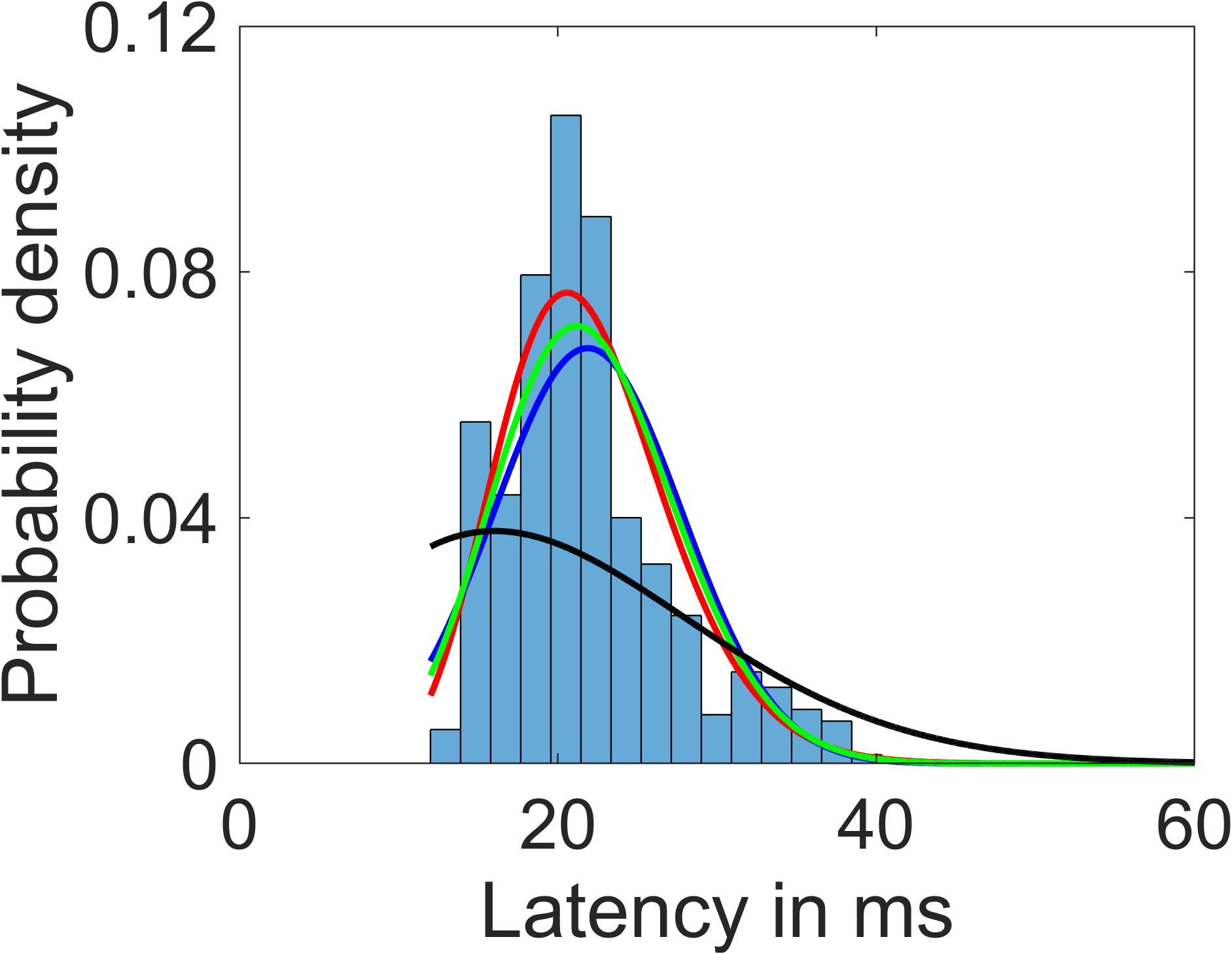}%
    \label{fig:China_Case_0}
  }
  \hfill
  \subfloat[]{%
    \includegraphics[width=0.23\linewidth]{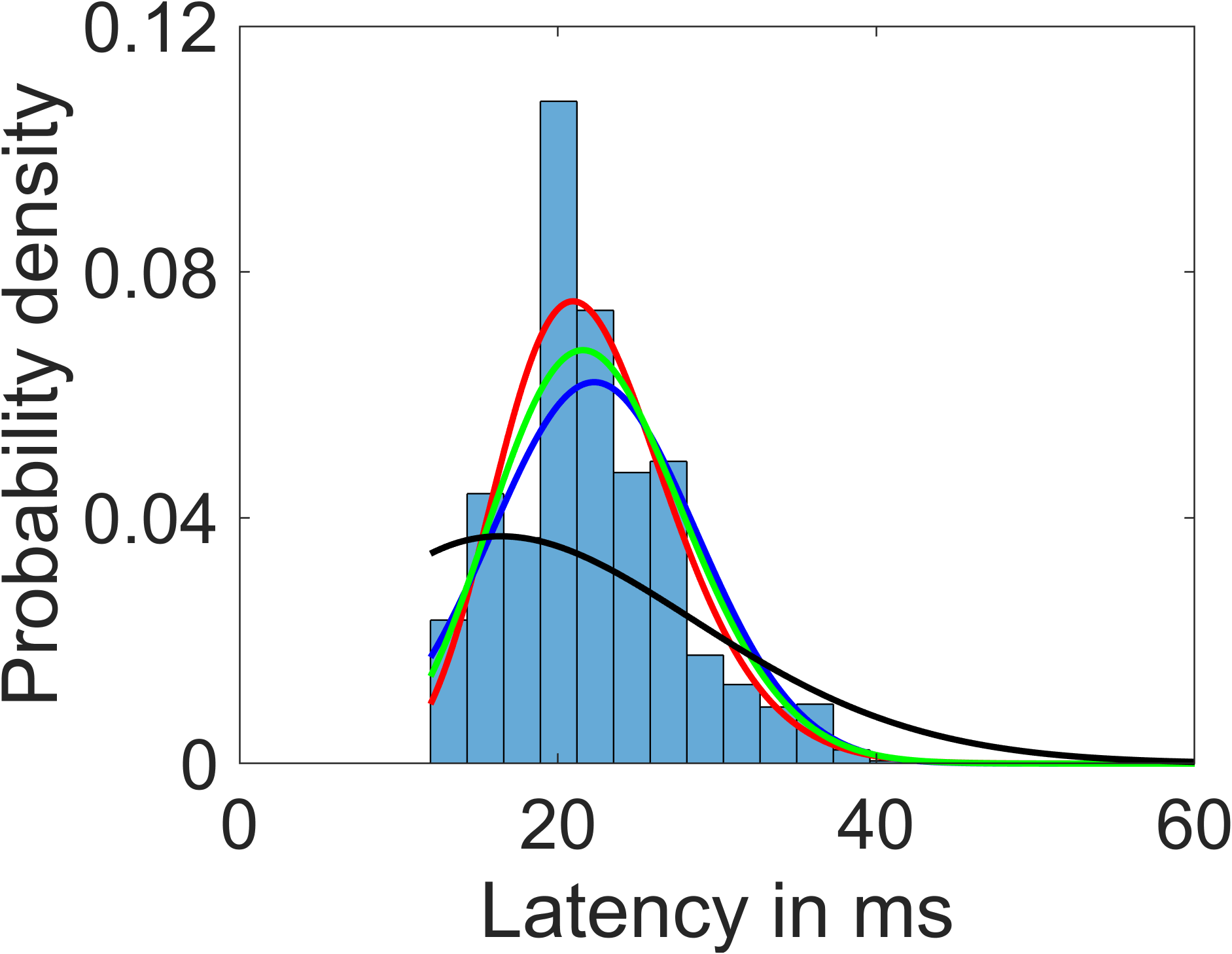}%
    \label{fig:China_Case_20}
  }
  \hfill
  \subfloat[]{%
    \includegraphics[width=0.23\linewidth]{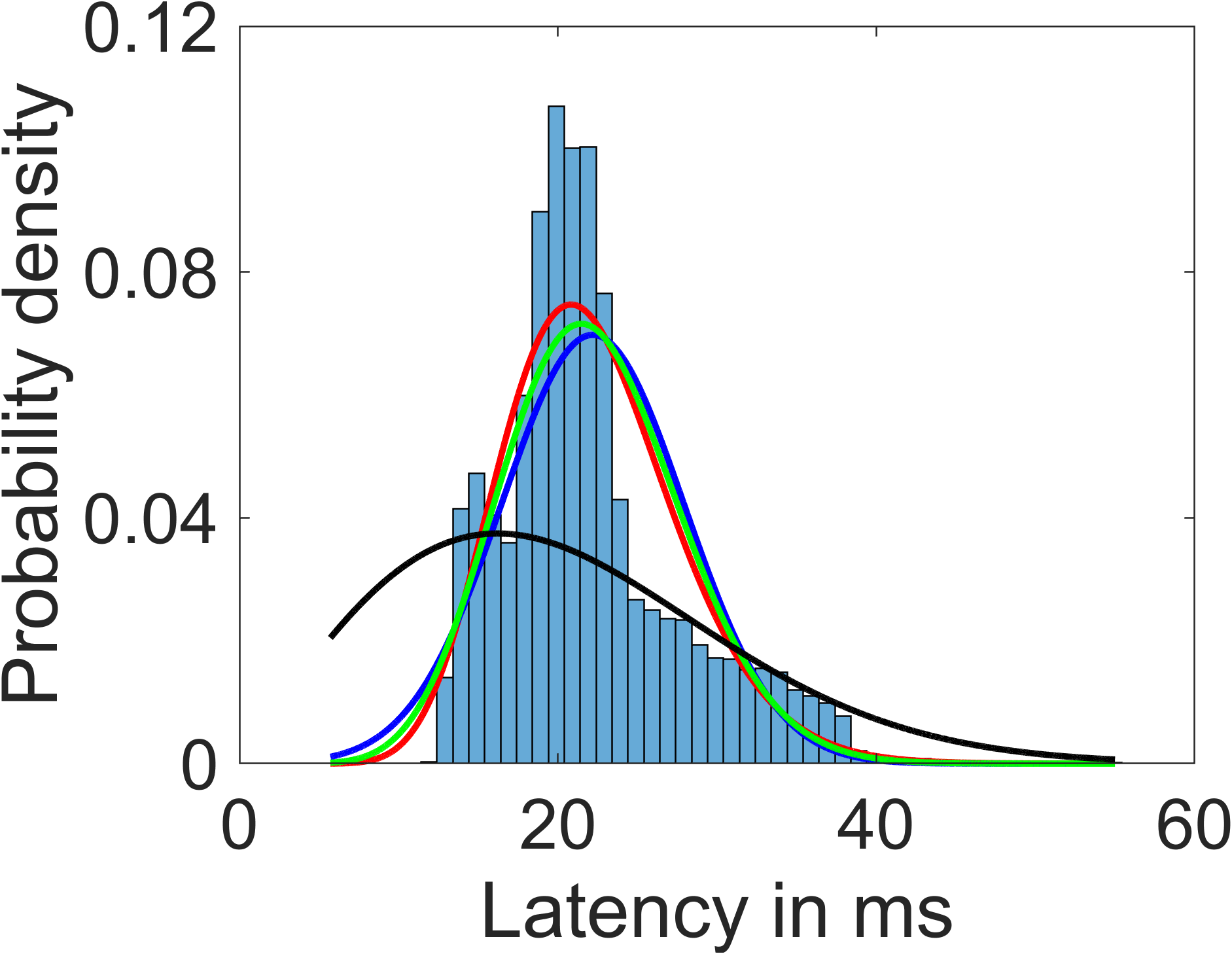}%
    \label{fig:China_Case_40}
  }
  \hfill
  \subfloat[]{%
    \includegraphics[width=0.23\linewidth]{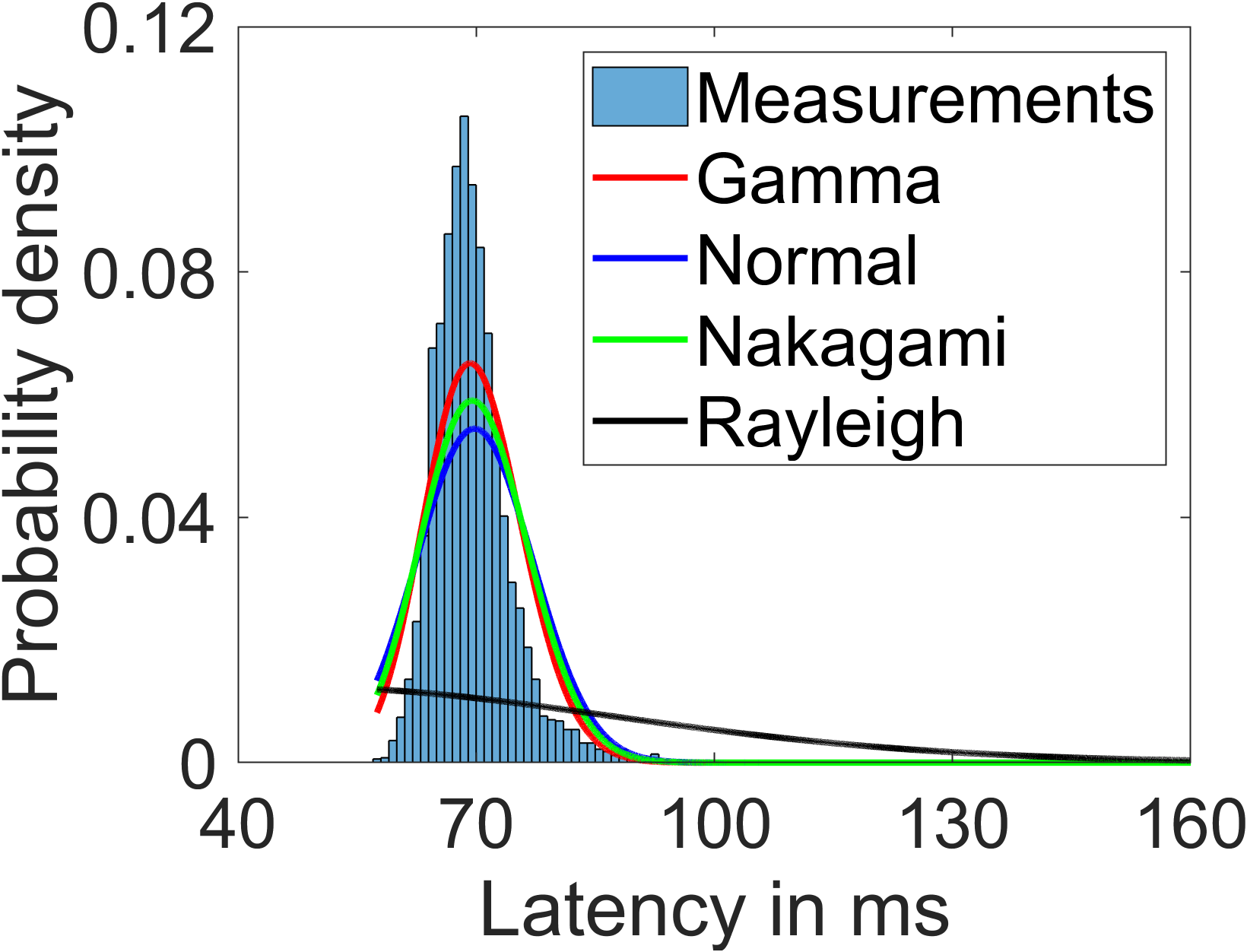}%
    \label{fig:Europe_Case}
  }
  \caption{Measured latency and fitted distributions of China cases with~\SI{0}{\kilo\meter\per\hour} (\ref{fig:China_Case_0}),~\SI{20}{\kilo\meter\per\hour} (\ref{fig:China_Case_20}),~\SI{40}{\kilo\meter\per\hour} (\ref{fig:China_Case_40}) and Europe cases (\ref{fig:Europe_Case}).}
  \label{fig:probability_China_Europe}
\end{figure*}

A measurement campaign was conducted at the ICV testing base of Tongji University, in Shanghai, China.~\added{Latency samples were recorded at a fixed interval of 50 ms, corresponding to the control-relevant timescale of cloud-controlled ICVs.} Tests were performed at stand still \SI{0}{\kilo\meter\per\hour}, as well as~\SI{20}{\kilo\meter\per\hour} and~\SI{40}{\kilo\meter\per\hour} traveling velocity. To enhance data reliability, each scenario was repeated at least ten times. Figure~\ref{fig:China_Case_0}--\ref{fig:China_Case_40} present the empirical latency distributions alongside fitted curves based on Gamma, Normal, Nakagami, and Rayleigh models. The quality of fit is evaluated using the sum of squared errors (SSE). As indicated in~\autoref{tab:SSE_China_use_case}, the \textbf{Gamma distribution} consistently achieved the lowest SSE across all test cases, demonstrating superior fitting performance.

\begin{table}[ht]
\caption{SSE between measured and fitted latency distributions}
\centering
\begin{tabular}{ccccc}
\toprule[2pt]
\multicolumn{1}{c}{} & \multicolumn{4}{c}{SSE} \\
\cmidrule(lr){2-5}
Velocity (km/h) & Normal & Nakagami & Rayleigh & \textbf{Gamma} \\
\midrule[1pt]
0  & 0.0265 & 0.0226 & 0.0688 & \textbf{0.0185} \\
20 & 0.0053 & 0.0047 & 0.0058 & \textbf{0.0038} \\
40 & 0.0091 & 0.0075 & 0.0080 & \textbf{0.0066} \\
\bottomrule[2pt]
\end{tabular}
\label{tab:SSE_China_use_case}
\vspace{2pt}
\noindent{\\\footnotesize{*\textbf{Bold entries} indicate the distribution with the lowest SSE, indicating the best fit to the measurements.}}
\end{table}

\begin{figure}[ht]
    \centering
    \includegraphics[width=\columnwidth]{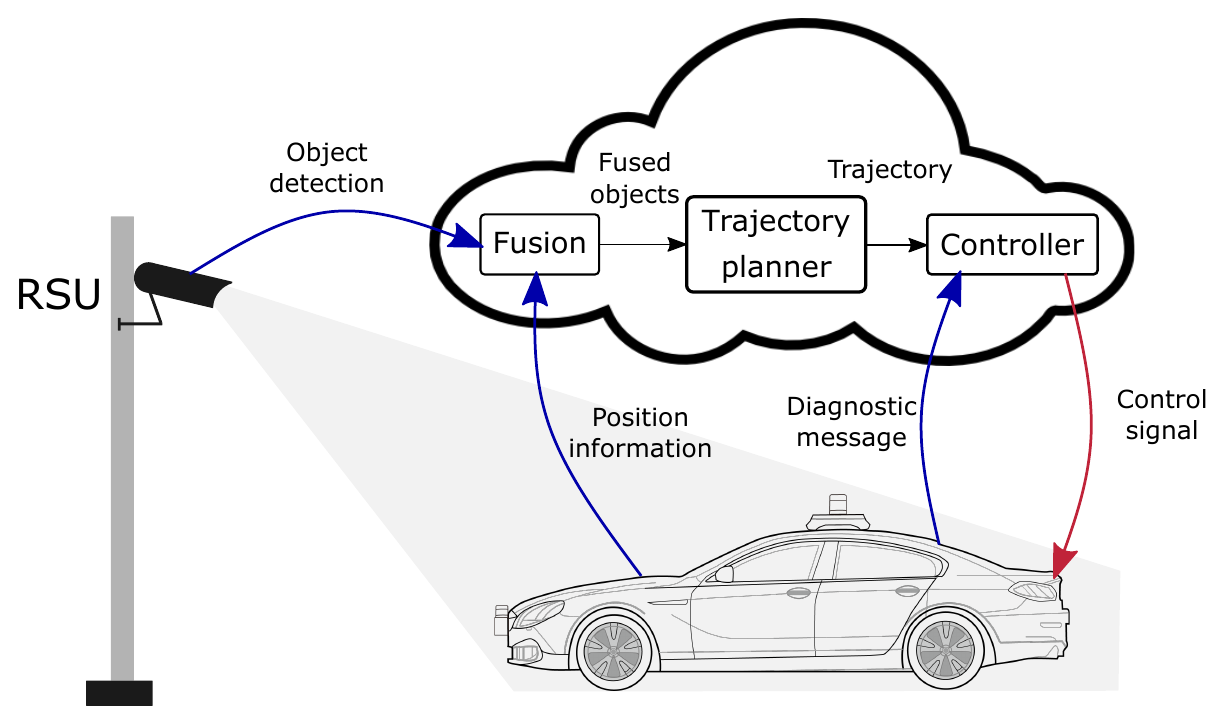}
    \caption{V2X communication solution in Europa use case.}
    \label{fig:fig_V2I}
\end{figure}

\subsubsection{Europe Use Case}

\subsubsection*{Overview of V2X Solution}
As shown in~\autoref{fig:fig_V2I}, the European V2X architecture comprises three main components: the RSU, test vehicle, and the central server (CS) that functions as a cloud-based processing unit. The RSU integrates sensors and an edge computer for object detection and tracking, forwarding the results to the CS. The test vehicle communicates its position to the CS and receives control signals for trajectory adjustment. A fusion algorithm within the CS combines inputs from both sources, which are then used by the trajectory planner to compute and return optimized trajectories. Further architectural details are provided in~\cite{reckenzaun2023transnational, tihanyi2021, tihanyi2021measurements}.

\subsubsection*{Probabilistic Model}
The measurement campaign was conducted at the ZalaZONE\footnote{\url{https://zalazone.hu/en/}} proving ground in Zalaegerszeg, Hungary~\cite{rovid2024digital}. The scenario involved a cloud-based control system reacting to a pedestrian dummy crossing as the test vehicle approached. The RSU monitored the scenario and relayed the data to the CS, simulating the cloud functionality. The CS then transmitted control signals to the vehicle to initiate an evasive maneuver.

The latency distribution for the European use case is based on five test trials, each lasting approximately~\SI{120}{\second} and involving the exchange of about 1200 messages exchanged between the test vehicle and the CS.~\added{The measurements were collected at a sampling interval of 100 ms.} Figure~\ref{fig:Europe_Case} shows the measured latency distribution along with fitted models, and~\autoref{tab:SSE_Europe_use_case} summarizes the SSE-based evaluation. Consistent with the China case results, the \textbf{Gamma distribution} again yields the best fit.

\begin{table}[ht]
\caption{SSE between measurements and fitted distributions}
    \centering
    \begin{tabular}{ccccc}
    \toprule[2pt]
        Model & Normal & Nakagami & Rayleigh & \textbf{Gamma} \\
        \midrule[1pt]
        SSE & 0.0080 & 0.0063 & 0.0258 & \textbf{0.0044} \\
        \bottomrule[2pt]
    \end{tabular}
    \label{tab:SSE_Europe_use_case}
    \vspace{2pt}
\noindent{\\\footnotesize{*\textbf{Bold entry} indicates the distribution with the lowest SSE, indicating the best fit to the measurements.}}
\end{table}

\added{Although the measurements in China and Hungary were collected at low speeds, they are also applied in highway scenarios, given the consistent 5G architecture and transmission paths across different speed levels. Prior study~\cite{zhang2025cicv5g} shows that the impact of vehicle speed on latency is moderate, supporting the use of these measurements for capturing latency trends under highway conditions.}

\subsubsection{Theoretical Model} To further validate the probabilistic model, a theoretical latency model is developed based on the structure of 5G communication networks. It incorporates queuing theory and retransmission mechanisms to analytically characterize latency. The derived latency distribution aligns with the \textbf{Gamma distribution}, thereby substantiating the empirical findings. The complete derivation process is detailed in Appendix~\ref{appendix:theoretical_latency_model}.

\subsubsection{\added{Adoption Strategy in Simulation}} \added{Empirical end-to-end (E2E) latency measurements from China and Hungary were fitted to a Gamma distribution with shape parameter $\alpha$ and scale parameter $\theta$.} \added{In the co-simulation, a Simulink block \texttt{gamrnd}}\footnote{\url{https://www.mathworks.com/help/stats/gamrnd.html}} \added{generates a random number from the fitted Gamma distribution with $\alpha$ and $\theta$ at each control cycle. The sampled value is converted from millisecond to second and fed into a Simulink \texttt{Variable Time Delay}}\footnote{\url{https://www.mathworks.com/help/simulink/slref/variabletimedelay.html}} \added{block, which applies the corresponding delay to the cloud-to-vehicle throttle and brake control signals.} \added{In this implementation, latency samples are drawn independently from the same distribution at each control cycle, following the common, i.e., they are modelled as independent and identically distributed.}

\section{Experiment}
\begin{figure*}[ht]
    \centering
    \begin{overpic}[width=\linewidth]{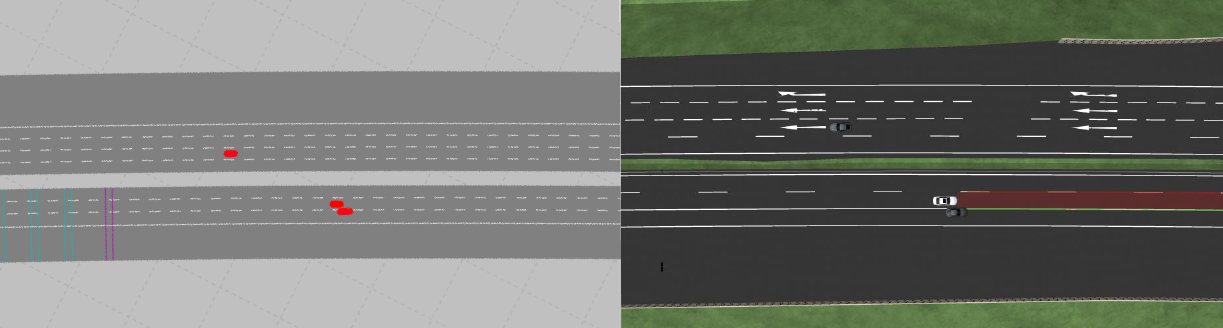}
        \put(3,23){\textcolor{white}{Vissim}}  
        
        \put(10,2){\textcolor{white}{\makecell[c]{Ego\\vehicle}}}  
        \put(15,3){\begin{tikzpicture}
            \draw[->, thick, white] (1, 0) -- (3.1, 1.3); 
        \end{tikzpicture}}
        
        \put(35,2){\textcolor{white}{\makecell[c]{Cut-in\\vehicle}}}  
        \put(29,2){\begin{tikzpicture}
            \draw[->, thick, white] (-3, 0) -- (-4, 1.3); 
        \end{tikzpicture}}

        \put(55,23){\textcolor{white}{CarMaker}}  
        
        \put(60,2){\textcolor{white}{\makecell[c]{Ego\\vehicle}}} 
        \put(65,3){\begin{tikzpicture}
            \draw[->, thick, white] (1, 0) -- (3, 1.3); 
        \end{tikzpicture}}

        \put(85,2){\textcolor{white}{\makecell[c]{Cut-in\\vehicle}}} 
        \put(79,2){\begin{tikzpicture}
            \draw[->, thick, white] (-3, 0) -- (-4, 1.3); 
        \end{tikzpicture}}
        
    \end{overpic}
    \caption{A snippet of a cut-in scenario in the co-simulation platform, where the cut-in vehicle is controlled by the PCM. The Left image shows the simulation in Vissim, while the right image presents the simulation in CarMaker.}
    \label{fig:platform_snippet}
\end{figure*}
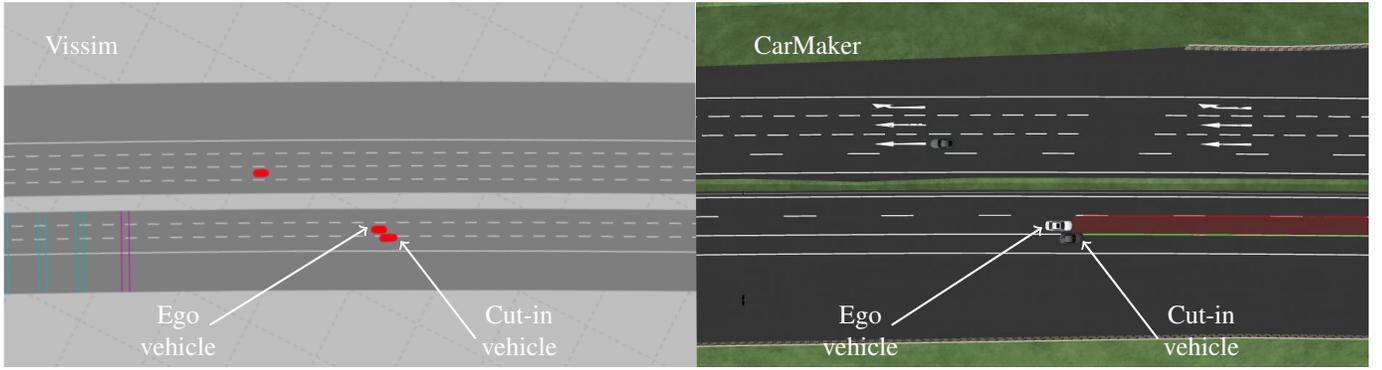

In Section~\ref{sec:method}, the methodology to construct the co-simulation platform is introduced, and a snippet of its implementation is shown in~\autoref{fig:platform_snippet}. In this section a simulation experiment is conducted and the results are analyzed to demonstrate the validity of this platform.

\subsection{Simulation Setup}
\label{sec:sim_setup}

\autoref{tab:sim_test_matrix} presents the simulation test matrix, comprising \replaced{eight}{six} conditions, defined by two PCM configurations (with and without PCM) and \replaced{four}{three} latency profiles: no latency (NL), China latency (CL), Hungary latency (HL)\added{, and abnormal latency (AL)}.

\begin{table}[ht]
\caption{Simulation Test Matrix}
\label{tab:sim_test_matrix}
    \centering
    \begin{tabular}{cccccc}
    \toprule[2pt]
            & NL & CL & HL & \added{AL}\\
        \midrule[1pt]
        Without PCM & $\surd$ & $\surd$ & $\surd$ & \added{$\surd$}\\
        \midrule[1pt]
        With PCM & $\surd$ & $\surd$ & $\surd$ & \added{$\surd$}\\
        \bottomrule[2pt]
    \end{tabular}
    \vspace{2pt}
\noindent{\\\footnotesize{*Each condition was simulated across five initial speeds (90, 100, 110, 120, 130 km/h) and three initial lane positions (left, center, right).}}
\end{table}

AL is designed as a conservative abnormal-latency stress-test scenario, rather than as a direct reproduction of the full empirical latency distribution. Specifically, it is derived from the combined China and Hungary latency measurements by selecting samples above the 99th percentile, fitting a normal distribution to these tail data, and drawing latency from the truncated distribution $\mathcal{TN}(\mu, \sigma^2; [q_{99}, m_{\max}])$, where $q_{99}$ is the 99th-percentile latency and $m_{\max}$ is the observed maximum. In this way, all simulated delays fall within $[q_{99}, m_{\max}]$, and the AL case intentionally shifts probability mass toward the worst-case regime to emulate prolonged congestion or outage conditions, while still remaining fully consistent with the measured extremes. Such a tail-focused construction, based on the 99th-percentile latency, is consistent with previous studies that evaluate V2X and cloud-based planning and control performance under extreme abnormal conditions~\cite{coll2022end}.

Each condition was evaluated under multiple initial configurations of the ego vehicle. These include five speeds and three lane positions, as summarized in~\autoref{tab:sim_test_matrix}. A simulation run refers to a complete run over the predefined scenario duration, which continues regardless of any collision occurrence.

\subsection{Safety Assessment}
The safety performance of the SUT is assessed using three complementary metrics: the collision rate (CR), the distance headway (DHW), and the post-encroachment time (PET). The CR measures the frequency of collisions to reflect overall safety. The DHW assesses the criticality of following scenarios, while the PET characterizes the criticality of cut-in scenarios encountered by the ego vehicle.

\subsubsection{Collision Rate}

\subsubsection*{Metric Definition}
The CR is computed cumulatively across all simulation runs by
\begin{equation}
\mathrm{CR} = \frac{\sum_{i=1}^{n} N_{\text{collision}}^{(i)}}{\sum_{i=1}^{n} D^{(i)}},
\label{eq:collision_rate}
\end{equation}
\noindent where \( N_{\text{collision}}^{(i)} \) denotes the number of collisions in the \( i \)-th simulation run, \( D^{(i)} \) is the driving distance in kilometers for the \( i \)-th run, and \( n \) represents the total number of runs.

\subsubsection*{Assessment Result}
\begin{figure}[ht]
    \centering
    \includegraphics[width=0.99\columnwidth]{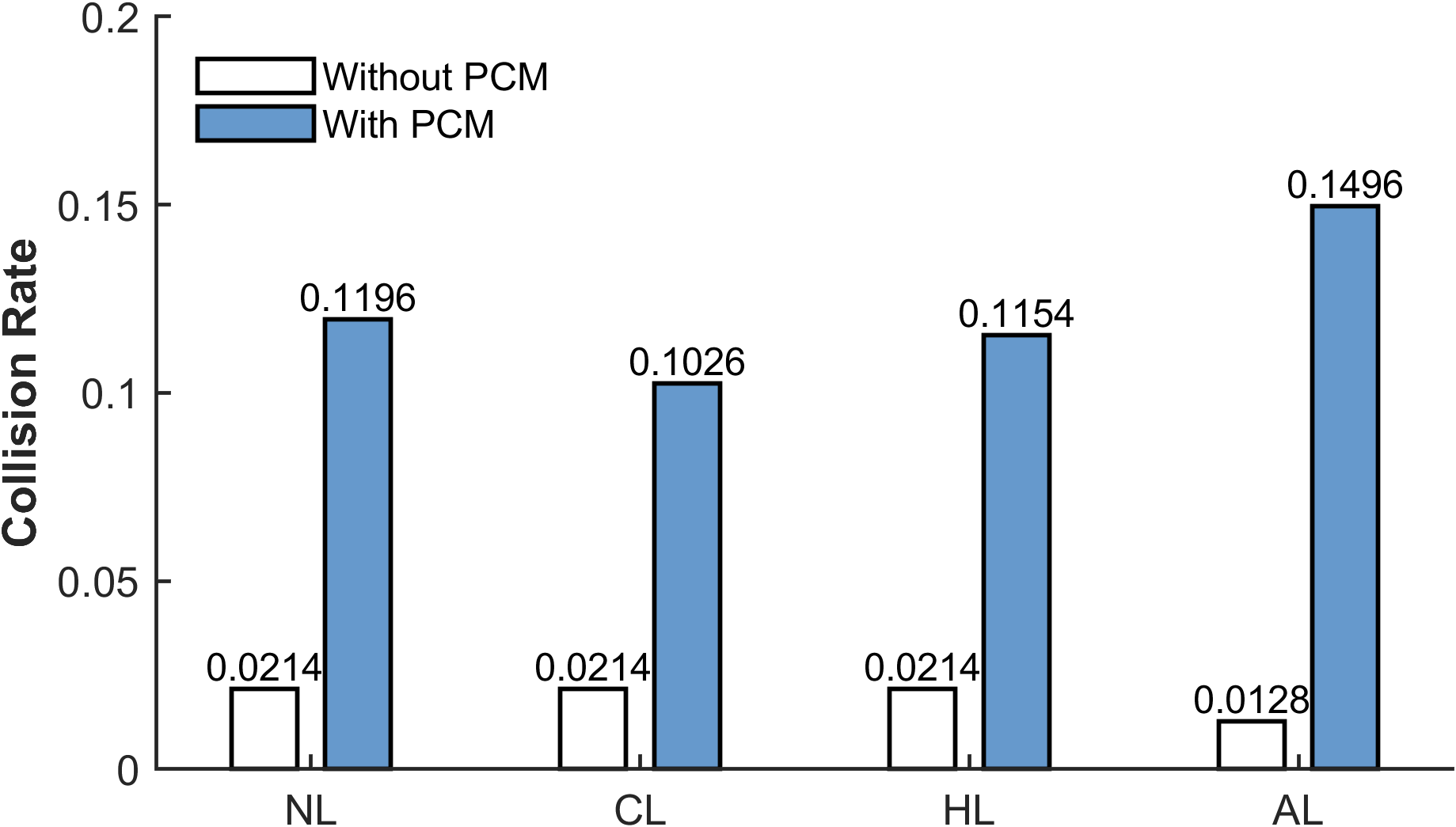}
    \caption{Collision rate of simulation test.}
    \label{fig:cr_sim_test}
\end{figure}

\autoref{fig:cr_sim_test} illustrates the CR under \replaced{eight}{six} test conditions. A markedly higher CR is observed when PCM is activated compared to tests without PCM across all latency settings. This suggests that the PCM effectively generates high-risk scenarios, thereby accelerating the exposure of the SUT to safety-critical scenarios.

When PCM is disabled, CR remains low and shows \minordel{no}\minoradd{negligible} variation across \minoradd{all} latency profiles\footnote{\minoradd{The slightly lower value under AL can be explained by two factors. First, when PCM is disabled, only a small number of collisions occur, so CR is sensitive to the presence or absence of a single collision. Second, AL makes the SUT slower and less aggressive due to the lag effect, which increases the following distance and reduces relative closing speed, thereby lowering CR.}}. When PCM is active, CR under NL, CL, and HL are comparable, whereas CR under AL is markedly higher\footnote{\minoradd{The slightly higher value in NL is explained by the fact that small latencies (CL/HL) introduce a millisecond‑level timing offset between the PCM-induced conflicts and the SUT's response. At highway speeds, a 50–100 ms offset corresponds to roughly 1–4 m, often turning a collision into a near miss. In NL, actions are tightly synchronized within the same simulation step, making exact spatiotemporal overlap more likely, and thus collisions are slightly more likely. The difference is small and within expected variability.}}. These findings indicate that \added{typical} latency cannot directly lead to collisions of ICVs in non-extreme scenarios, \added{while AL can substantially increase collision occurrences.} \replaced{This observation}{which} aligns with previous work~\cite{zhang2024cloud}.


\subsubsection{Distance Headway}

\subsubsection*{Metric Definition}
In following scenarios, the DHW denotes the Euclidean distance between the ego and the lead vehicle at each time step, defined~\replaced{by~\autoref{eq:dst}.}{as }\deleted{$\mathrm{DHW}(t) = \left\| \mathbf{p}_{\text{lead}}(t) - \mathbf{p}_{\text{ego}}(t) \right\|,$}
\deleted{where \( \mathbf{p}_{\text{lead}}(t) \) and \( \mathbf{p}_{\text{ego}}(t) \) denote the positions of the lead and ego vehicles at time \( t \), respectively.} Based on~\cite{junietz2018pegasus}, a highway following scenario is considered critical when the DHW falls below~\SI{50}{\meter}. Then, the frequency of a critical following scenario $f_{\text{crit}}^{\text{DHW}}$ is computed by
\begin{equation}
    f_{\text{crit}}^{\text{DHW}} = \frac{N_{\text{DHW}<50}}{N_{\text{total}}},
\end{equation}
\noindent where $N_{\text{DHW}<50}$ is the number of time steps with DHW below~\SI{50}{\meter}, and $N_{\text{total}}$ is the total number of time steps during the simulation.

\subsubsection*{Assessment Result}
\autoref{tab:freq_dhw} presents the frequency of critical following scenarios $f_{\text{crit}}^{\text{DHW}}$ under each test condition. With PCM activated, the frequency increases by \replaced{2.9 to 3.5}{1.2 to 1.5} times compared to when PCM is disabled, indicating that PCM substantially elevates the criticality of following scenarios.

\begin{table}[ht]
\caption{Frequency of Critical Following Scenarios}
\label{tab:freq_dhw}
\centering
\small
\begin{tabular}{lcccc}
    \toprule[2pt]
    & NL & CL & HL & \added{AL}\\
    \midrule[1pt]
    Without PCM & \replaced{0.1016}{0.199} & \replaced{0.1014}{0.205} & \replaced{0.1002}{0.199} & \added{0.1129}\\
    With PCM    & \replaced{0.4423}{0.484} & \replaced{0.4572}{0.465} & \replaced{0.4625}{0.508} & \added{0.4463}\\
    \textbf{Increase} & \textbf{\replaced{+335.2\%}{+143.2\%}} & \textbf{\replaced{351.0\%}{+126.8\%}} & \textbf{\replaced{+361.7\%}{+155.3\%}} & \added{\textbf{+295.5\%}}\\
    \bottomrule[2pt]
\end{tabular}
\vspace{2pt}
\noindent{\\\footnotesize{*\textbf{Bold values} indicate the percentage increase in the frequency of critical following scenarios caused by PCM activation.}}
\end{table}

\autoref{tab:freq_diff} reports the relative changes in frequency of critical following scenarios compared to the NL condition. Without PCM, CL and HL show no increase, whereas AL causes a clear rise\footnote{\minoradd{Without PCM, the slightly higher value in NL compared to CL/HL. The reason is the same as noted in Footnote 8. This minor effect does not alter conclusions.}}. With PCM, all latency conditions (CL, HL, AL) raise the frequency. The small value under AL with PCM (+0.90\%) reflects the markedly larger total number of following scenarios in this setting; despite this, the absolute frequency of critical following scenarios is the highest \minoradd{(13883)} among all conditions\minoradd{, as shown in~\autoref{tab:follow_crit_total}}.

\begin{table}[ht]
\caption{Relative Change in Critical Following Scenario Frequency (NL vs. Others)}
\label{tab:freq_diff}
\centering
\small
\begin{tabular}{lcc}
    \toprule[2pt]
    Comparison & Without PCM & With PCM \\
    \midrule[1pt]
    NL vs. CL & \replaced{-0.26\%}{\textbf{+3.0\%}} & \replaced{\textbf{+3.37\%}}{-3.9\%} \\
    NL vs. HL & \replaced{-1.42\%}{0.0\%} & \replaced{\textbf{+4.57\%}}{\textbf{+5.0\%}} \\
    \added{NL vs. AL} & \added{\textbf{+11.02\%}} &  \added{\textbf{+0.90\%}}\\
    \bottomrule[2pt]
\end{tabular}
\vspace{2pt}
\noindent{\\\footnotesize{*\textbf{Bold values} indicate the percentage increase in critical following scenario frequency, computed by comparing CL, \replaced{HL, and AL}{and HL} against the NL baseline.}}
\end{table}

\begin{table}[ht]
\caption{Absolute Frequency of Following Scenarios}
\label{tab:follow_crit_total}
\centering
\small
\begin{tabular}{llcccc}
    \toprule[2pt]
    & & \minoradd{NL} & \minoradd{CL} & \minoradd{HL} & \minoradd{AL}\\
    \midrule[1pt]
    \multirow{2}{*}{\minoradd{Without PCM}} & \minoradd{Critical} & \minoradd{2060} & \minoradd{2084} & \minoradd{2045} & \minoradd{2647} \\
    & \minoradd{Total} & \minoradd{20269} & \minoradd{20559} & \minoradd{20412} & \minoradd{23459} \\
    \midrule[1pt]
    \multirow{2}{*}{\minoradd{With PCM}} & \minoradd{Critical} & \minoradd{11712} & \minoradd{11533} & \minoradd{12392} & \minoradd{\textbf{13883}} \\
    & \minoradd{Total} & \minoradd{26480} & \minoradd{25225} & \minoradd{26792} & \minoradd{31109} \\
    \bottomrule[2pt]
\end{tabular}
\vspace{2pt}
\noindent{\\\footnotesize{*\textbf{Bold value} indicates the most critical following scenario counts in AL.}}
\end{table}

\subsubsection{Post-Encroachment Time} 

\subsubsection*{Metric Definition}
In cut-in scenarios, the PET quantifies the temporal gap between the completion of a cut-in vehicle's lane change and the ego vehicle reaching the same location, with a defined spatial tolerance $\delta$. PET is computed by
\begin{equation}
\text{PET} = t_{\text{ego}} - t_{\text{cut}},
\end{equation}
\noindent where \( t_{\text{cut}} \) is the time when the cut-in vehicle completes the maneuver and \( t_{\text{ego}} \) is the earliest time when the ego vehicle reaches the same location (within $\delta$). Let \( \mathbf{p}_{\text{cut}}^* \) be the position of the cut-in vehicle at time \( t_{\text{cut}} \), then
\begin{equation}
t_{\text{ego}} = \min \left\{ t \geq t_{\text{cut}} \;\middle|\; \left\| \mathbf{p}_{\text{ego}}(t) - \mathbf{p}_{\text{cut}}^* \right\| < \delta \right\}.
\end{equation}
Following~\cite{VARHELYI1998731}, a cut-in scenario is regarded as critical if the PET is below~\SI{1}{\second}.
\deleted{Then, the frequency of critical cut-in scenarios $f_{\text{crit}}^{\text{PET}}$ is computed by $f_{\text{crit}}^{\text{PET}} = \frac{N_{\text{PET} < 1s}}{N_{\text{cut-in}}}$,} 
\added{Then, the critical cut-in scenarios rate (CCSR) is computed by}
\begin{equation}
\text{CCSR} = \frac{\sum_{i=1}^{n} N_{\text{PET} < 1s}^{(i)}}{\sum_{i=1}^{n} D^{(i)}},
\label{eq:critical_pet_rate}
\end{equation}
\noindent \replaced{where $N_{\text{PET} < 1s}^{(i)}$ denotes the number of critical cut-in scenarios in the \( i \)-th simulation run, \( D^{(i)} \) is the driving distance in kilometers for the \( i \)-th run, and \( n \) represents the total number of runs.}{where $N_{\text{PET} < 1s}$ is the number of cut-in scenarios with PET below 1s, and $N_{\text{cut-in}}$ is the total number of cut-in scenarios during the simulation.}

\subsubsection*{Assessment Result}
\autoref{tab:freq_pet} summarizes the \replaced{CCSR}{frequency of critical cut-in scenarios $f_{\text{crit}}^{\text{PET}}$} under different test conditions. A marked increase is observed when PCM is activated, indicating that the PCM effectively generates high-risk cut-in scenarios on highways.

\begin{table}[ht]
\caption{\replaced{Critical cut-in scenario rate\label{tab:freq_pet}}{Frequency of Critical Cut-in Scenarios}}
\centering
\small
\begin{tabular}{lcccc}
    \toprule[2pt]
    & NL & CL & HL & \added{AL}\\
    \midrule[1pt]
    Without PCM & \replaced{1.282}{0.132} & \replaced{0.855}{0.143} & \replaced{1.709}{0.132} & \added{1.282}\\
    With PCM    & \replaced{17.947}{0.343} & \replaced{18.802}{0.312} & \replaced{20.738}{0.320} & \added{21.793}\\
    \textbf{Increase} & \textbf{\replaced{+1300\%}{+159.8\%}} & \textbf{\replaced{+2100\%}{+118.2\%}} & \textbf{\replaced{+1113\%}{+142.4\%}} & \minoradd{\textbf{+1600\%}}\\
    \bottomrule[2pt]
\end{tabular}
\vspace{2pt}
\noindent{\\\footnotesize{*\textbf{Bold values} indicate the percentage increase in the \replaced{critical cut-in scenario rate}{frequency of critical cut-in scenarios} caused by PCM activation.}}
\end{table}

\added{\autoref{tab:freq_pet_diff} reports the relative changes in CCSR. Without PCM the effects are mixed and no consistent impact is observed. With PCM enabled, all latency profiles increase CCSR; AL shows the largest increase, while CL and HL yields smaller gains. This observation suggests that latency increases critical cut-in scenarios only in extreme driving conditions.} \deleted{reports the relative changes in critical cut-in scenario frequency. The results suggest that CL and HL exhibit inconsistent effects on enhancing criticality of cut-in scenarios, regardless of PCM activation. This indicates that the effect of latency on ICV safety performance cannot be substantiated through enhancing criticality in cut-in scenarios.}

\begin{table}[ht]
\caption{Relative Change in Critical Cut-in Scenario \replaced{Rate\label{tab:freq_pet_diff}}{Frequency} (NL vs. Others)}
\centering
\small
\begin{tabular}{lcc}
    \toprule[2pt]
    Comparison & Without PCM  & With PCM  \\
    \midrule[1pt]
    NL vs. CL & \replaced{-33.33\%}{\textbf{+8.3\%}} & \replaced{\textbf{+4.76\%}}{-9.0\%} \\
    NL vs. HL & \replaced{+33.33\%}{0.0\%}        & \replaced{\textbf{+15.55\%}}{-6.7\%} \\
    \added{NL vs. AL} & \added{0\%}        & \added{\textbf{+21.43\%}} \\
    \bottomrule[2pt]
\end{tabular}
\vspace{2pt}
\noindent{\\\footnotesize{*\textbf{Bold values} indicate the percentage increase in critical cut-in scenario \replaced{rate}{frequency}, computed by comparing CL\replaced{, HL and AL}{and HL} against the NL baseline.}}
\end{table}

\subsection{Comfort Assessment}

\subsubsection{Metric Definition}
The comfort metric is defined as the total energy of longitudinal acceleration within the ISO 2631-defined sensitive frequency band (0.5--10~Hz)~\cite{ISO_2001}, obtained via discrete Fourier transform (DFT) of the absolute acceleration signal. The power spectral density $P(f)$ is calculated by
\begin{equation}
P(f) = \frac{|\mathcal{F}(a(t))|^2}{N}, 
\label{eq:psd}
\end{equation}

\noindent where $a(t)$ is the absolute longitudinal acceleration, $\mathcal{F}(a(t))$ is its Fourier transform, $N$ is the number of samples. The total power in sensitive frequency band  \( E_{\text{sens}} \) is then computed by summing $P(f)$ over the range of $f \in [0.5, 10]$~Hz
\begin{equation}
E_{\text{sens}} = \sum_{f = 0.5~\text{Hz}}^{10~\text{Hz}} P(f).
\label{eq:sum_power}
\end{equation}

\subsubsection{Assessment Result}
\autoref{tab:power_longitudinal} presents the total power within sensitive frequency across all test conditions. The results show that PCM activation leads to a substantial increase in power compared to the no-PCM case. This indicates that PCM significantly degrades ride comfort.

\begin{table}[ht]
\caption{Total Power within Sensitive Frequency (\SI{}{(m/s^2)^2})}
\label{tab:power_longitudinal}
\centering
\small
\begin{tabular}{lcccc}
    \toprule[2pt]
    & NL & CL & HL & \added{AL}\\
    \midrule[1pt]
    Without PCM & 6566.51  & 6796.71  & 7827.16  & \added{11803.12}\\
    With PCM    & 20522.47 & 21412.07 & 24380.87 & \added{31415.90}\\
    \textbf{Increase}   & \textbf{+212.6\%} & \textbf{+214.9\%} & \textbf{+211.4\%} & \added{\textbf{+166.1\%}}\\
    \bottomrule[2pt]
\end{tabular}
\vspace{2pt}
\noindent{\\\footnotesize{*\textbf{Bold values} indicate the percentage increase in total power in sensitive frequency caused by PCM activation.}}
\end{table}

\autoref{tab:power_diff} presents the relative change in total power within sensitive frequency. The results show that latency increases this power to varying extents under both PCM-on and PCM-off conditions. These findings suggest that latency tends to reduce ride comfort by amplifying longitudinal acceleration in the sensitive frequency range.

\begin{table}[ht]
\caption{Relative Change in Total Power in Sensitive Frequency (NL vs. Others)}
\label{tab:power_diff}
\centering
\small
\begin{tabular}{lcc}
    \toprule[2pt]
    Comparison & Without PCM & With PCM \\
    \midrule[1pt]
    NL vs. CL & \textbf{+3.5\%}   & \textbf{+4.3\%} \\
    NL vs. HL & \textbf{+19.2\%}  & \textbf{+18.8\%} \\
    \added{NL vs. AL} & \added{\textbf{+79.7\%}}  & \added{\textbf{+53.1\%}} \\
    \bottomrule[2pt]
\end{tabular}
\vspace{2pt}
\noindent{\\\footnotesize{*\textbf{Bold values} indicate the percentage increase in total power within sensitive frequency, computed by comparing CL\replaced{, HL and AL}{and HL} against the NL baseline.}}
\end{table}

In summary, the following conclusions and findings are obtained through the experiment:
\begin{enumerate}
    \item The proposed co-simulation platform, through the use of the PCM, effectively increases the exposure of the SUT to safety-critical scenarios, particularly by generating more frequent collisions (up to \replaced{11 times}{fivefold}) and enhancing the criticality of following and cut-in scenarios by \replaced{3}{1.2} to \replaced{13}{1.5} times. This demonstrates the platform's capability in accelerating safety validation.
    \item \added{The impact of latency on ICV safety performance is conditional on both the driving situations and latency types. Normal latency (CL, HL) does not increase collisions, yet it amplifies criticality in extreme scenarios (PCM enabled). Conversely, AL induces collisions and elevates driving risk.}\deleted{V2C communication latency does not significantly affect the safety performance of ICVs. It neither increases the collision rate nor intensifies the criticality of following or cut-in scenarios. However, it consistently reduces ride comfort by amplifying longitudinal acceleration within the sensitive frequency range, leading to power increases of up to 20\%. This finding is consistent with the observation reported in~\mbox{\cite{zhang2024cloud}.}}
    \item \added{All types of latency consistently reduces ride comfort by amplifying longitudinal acceleration within the sensitive frequency range, leading to power increases of up to 20\%. This finding is consistent with the observation reported in~\cite{zhang2024cloud}.}
\end{enumerate}

\section{Conclusion}

This study introduces a co-simulation platform that combines a detailed vehicle dynamics model, a realistic traffic environment based on real-world measurements, strategic BGV control, and a V2C latency model for assessing ICVs. The platform’s validity is demonstrated through testing an exemplary SUT across \replaced{eight}{six} conditions, combining PCM activation with \replaced{four}{three} latency settings. The simulation results indicate that the platform can generate high-risk driving environments, resulting in approximately a fivefold increase in collisions\replaced{, a 3-fold increase in critical following scenarios, and up to a 21-fold increase in the critical cut-in scenario rate when PCM is activated.}{and 1.2 to 1.5 times more safety-critical following and cut-in scenarios.} Moreover, the results reveal that V2C communication latency has a primary impact on ICV comfort\replaced{, while under extreme driving conditions it also amplifies safety-critical scenarios}{rather than safety performance}, which is consistent with findings reported in~\cite{zhang2024cloud}.

A practical limitation of the platform is slower simulation speed when the PCM is activated to control BGVs, as frequent data exchange introduces delays.~\added{While PCM activation increases simulation time by about four- to fivefold, it also highlights opportunities for future improvement. Possible directions include exploring event-triggered updates schemes to reduce synchronization load, developing more efficient data-exchange interfaces that lower coding complexity while ensuring interoperability, and deployment on high-performance computing platforms (multi-core CPU servers) to support large-scale testing.} \deleted{Improving the communication interface or running the controller in parallel can speed up the simulation.}

In future research, critical scenarios formatted with OpenSCENARIO, generated using the tool developed in~\cite{10588843}, will be incorporated to enable scenario-based testing. Additionally, Vehicle-to-Infrastructure (V2I) cooperative perception will be incorporated to address perception limitations and information silos. Furthermore, we will systematically compare alternative tail models and percentile thresholds for the AL scenario, including temporal correlation effects, to further refine the representation of abnormal network conditions in the proposed co-simulation platform.


{\appendices
\section{Vehicle Dynamics Model}
\label{appendix:vehicle_dynamic_model}

\subsection{Parameter identification} \label{appx:par_identi}

The required model parameters and characteristics are outlined in~\autoref{tab:req_params}. In the following, the main topics of the identification process are summarized.

\begin{itemize}
    \item \textbf{Mass and center of gravity (COG):} Parameters were experimentally identified by laboratory measurements, applying the method of~\cite{kollreider2009}. The total mass as well as the position of the COG in longitudinal and lateral direction were determined by wheel load measurements on scales on a flat surface. The hight of the COG (vertical direction) was determined by lifting one axle of the vehicle.
    
    \item \textbf{Moments of inertia (MOI):} Measurement of the inertia tensor of the specific test vehicle was not possible. However, data of measured MOI of a reference vehicle was available. Therefore, the MOI regarding the pitch, roll and yaw motions were approximated in a simple but effective approach. The respective measured values of the reference vehicle $I^\text{meas}_\text{ref}$ were compared to the value of an idealized rigid cuboid $I^\text{cub}_\text{ref}$, approximating the overall size of the reference vehicle, by
    \begin{equation} \label{eq:correction_coefficient}
    i_{\text{corr}} = \frac{I^\text{meas}_\text{ref}}{I^\text{cub}_\text{ref}}\,.
    \end{equation}
    Additionally, it was assumed that the ratio described by the dimensionless correction factor $i_{\text{corr}}$ is similar for the test vehicle. Hence, the MOI of the test vehicle $I_\text{ego}^\text{approx}$ were approximated by
    \begin{equation} \label{eq:moi_correction}
    I_\text{ego}^\text{approx} = I^\text{cub}_\text{ego}\,i_\text{corr}\,,
    \end{equation}
    where $I^\text{cub}_\text{ego}$ represents the MOI of a cuboid with the size of the test vehicle.
    
    \item \textbf{Spring characteristics:} Suspension spring characteristics and stiffness were experimentally identified by laboratory measurements, applying the method of~\cite{leichtfried2015}. The test vehicle was loaded with additional weights in certain stages, which lead to increasing wheel loads and vertical suspension travel. The wheel loads were measured by scales and the vertical suspension travel by cable-extension transducers positioned at each wheel. Consequently, spring characteristics were determined and the stiffness identified. This pragmatic and effective approach leads to sufficiently accurate results, avoiding cumbersome and expensive measurements on test rigs.
    
    \item \textbf{Damping characteristics:} Measurement of suspension damping characteristics was not possible. Additionally, the test vehicle is equipped with an active damping system, where no data was available. Therefore, suspension damping properties $c_\text{S}$ were approximated based on the dimensionless damping ratio $\zeta$, assuming simple vertical dynamics of a quarter vehicle (one mass oscillator) by
    \begin{equation} \label{eq:damping_ratio}
        \zeta = \frac{c_\text{S}}{2\,\sqrt{k_{\text{S}}\,m}}\,,
    \end{equation}
    where $m$ denotes a quarter of the total vehicle mass and $k_{\text{S}}$ indicates the suspension spring stiffness. A sportive design was assumed and different damping properties for compression and rebound cycles, resulting in overall nonlinear damping characteristics.

    \item \textbf{Anti-roll bar characteristics:} The test vehicle is equipped with actively controlled anti-roll bar characteristics on the front as well as on the rear axle. No information about the active system was available. Therefore, anti-roll bar characteristics were modeled by pragmatic approaches, resulting in a vehicle velocity and lateral acceleration dependent anti-roll bar stiffness. These, characteristics were identified using measurement data of closed-track vehicle dynamics testing on a proving ground.
    
    \item \textbf{Steering ratio:} The test vehicle is equipped with active front and rear wheel steering. In a first step, steering characteristics and the steering ratio were experimentally identified by laboratory measurements in standstill. Measurements were carried out with a measurement steering wheel and wheel protractors. Subsequently, the identified steering ratio in standstill was adjusted for the active, velocity dependent steering system, based on closed-track vehicle dynamics testing on a proving ground. 

    \item \textbf{Tire force characteristics:} Data of the test vehicle's tires was not available. Therefore, measured tire force characteristics of a similar tire, obtained by tire testing on an industrial flat track tire test rig, were used. Similar tire properties were assumed.
    
    \item \textbf{Powertrain characteristics:} Power and torque characteristics, as well as gear ratios, were obtained from the data sheet of the vehicle.
\end{itemize}

\subsection{Modeling of subsystems and implementation} \label{appx:modeling}

The adaptable multi-body framework of CarMaker was used to consider the actively controlled subsystems of the test vehicle, in particular the active anti-roll bars as well as the front and rear wheel steering systems. These systems were modeled in MATLAB/Simulink, based on measurement data from closed-track vehicle dynamics measurements and implemented as Functional Mock-up Units (FMUs) in CarMaker. Tire force transmission was modeled by Pacejka's Magic Formula tire model, which is directly available in CarMaker. For the remaining subsystems and characteristics, such as brake system and suspension kinematics, generic templates of CarMaker were used.

\subsection{Evaluation and experimental validation} \label{appx:validation}

The performance of the vehicle model as well as the plausibility of simulation results were evaluated and validated by means of vehicle dynamics measurement data. Tests were conducted on a closed-track proving ground, including steady-state cornering on a~\SI{90}{\meter} circular track, sinusoidal steering maneuvers at~\SI{55}{\kilo\meter\per\hour} and~\SI{80}{\kilo\meter\per\hour} as well as different acceleration and braking maneuvers, see~\cite{reichmann2024}.

Measurement data was collected using a Genesys ADMA\footnote{\url{https://genesys-offenburg.de/adma-g/}} device, which recorded speeds, accelerations, position, and attitude angles. Additionally, the steering wheel angle was measured by a measurement steering wheel.

Subsequently, measurements and simulation results were compared by using measured steering angle and vehicle speed as inputs for the model, to replicate the closed-track tests. In~\autoref{tab:model_vali}, the calculated Root Mean Square Errors (RMSEs) between measurement and simulation results of relevant quantities are listed, such as longitudinal velocity $v_x$, yaw rate $\omega_z$, roll angle $\phi$ as well as longitudinal and lateral accelerations $a_x$ and $a_y$, respectively.

Although there are higher deviations of the yaw rate, especially during sinusoidal steering, and further validations are recommended, the overall results indicate that the vehicle model is valid for the investigations conducted in the present work.

\begin{table}[ht]
\caption{RMSE between simulation results and measurements, based on~\cite{reichmann2024}.}\label{tab:model_vali}
\centering
\begin{tabular}{llll}
\toprule[2pt]
RMSE & \makecell[l]{Steady-state\\cornering} & \makecell[l]{Sinusoidal\\steering} & \makecell[l]{Acceleration\\maneuver}\\
\midrule[1pt]
\(v_x\)~(\SI{}{\kilo\meter\per\hour}) & 0.48 & 0.38 & 0.58\\
\(a_x\)~(\SI{}{\meter\per\second\squared}) & 0.31 & 0.68 & 0.55\\
\(a_y\)~(\SI{}{\meter\per\second\squared}) & 0.23 & 0.56 & 0.31\\
\(\omega_z\)~(\SI{}{\degree\per\second}) & 0.93 & 3.49 & 1.01\\
\(\phi\)~(\SI{}{\degree}) & 0.30 & 0.34 & NA\\
\bottomrule[2pt]
\end{tabular}
\end{table}

\section{Theoretical Latency Model}
\label{appendix:theoretical_latency_model}
The theoretical latency model is derived and validated against the actual communication latency depicted in~\autoref{fig:China_Europe_architecture}. The end-to-end (E2E) 5G communication latency~\(l_{\text{E2E}}\) includes the radio access network (RAN)~\(l_{\text{radio}}\), transport network (TN)~\(l_{\text{TN}}\), core network (CN) latency~\(l_{\text{CN}}\), the latency between the CN' UPF node and the AS~\(l_{\text{UPF-AS}}\), and the AS processing latency~\(l_{\text{AS}}\). This is expressed as
\begin{equation} 
\label{eq:end2end_latency}
    l_{\text{E2E}} = l_{\text{radio}} + l_{\text{TN}} + l_{\text{CN}} + l_{\text{UPF-AS}} + l_{\text{AS}}\,.
\end{equation}
Each of these components includes both uplink and downlink latencies. 

In practice, besides considering retransmissions in the RAN, the latencies of other nodes can be simplified to the sum of queuing latency and transmission latency~\cite{9964110}. Therefore, queuing theory is utilized for detailed RAN modeling, while the latencies at other nodes are similar. RAN latency can be formulated as
\begin{equation} \label{eq: RAN latency_1}
    l_{\text{radio}} = \tau_{\text{radio}} + \tau_{\text{HARQ}} + N\,(\tau^\prime_{\text{radio}} + \tau_{\text{HARQ}})\,,
\end{equation}
with \(N={0,1,..., N_{\text{max}}}\), and where \(\tau_{\text{radio}}\) and \(\tau^\prime_{\text{radio}}\) are the transmission and retransmission latency between the onboard device and the gNB, respectively. The latency caused by the hybrid automatic repeat request (HARQ) retransmission mechanism is represented by~\(\tau_{\text{HARQ}}\), and \(N\) denotes the number of retransmissions. According to the internet protocol (IP) layer data scheduling mechanism,~\autoref{eq: RAN latency_1} can be reformulated as
\begin{equation} \label{eq: RAN latency}
    l_{\text{radio}} = \tau_{\text{sch}} + \tau_{\text{data}} + \tau_{\text{HARQ}} + N\,(\tau^\prime_{\text{sch}} + \tau_{\text{data}} + \tau_{\text{HARQ}})\,,
\end{equation}
where \(\tau_{\text{sch}}\) represents the resource scheduling latency, \(\tau_{\text{data}}\) denotes the data transmission latency, and \(\tau_{\text{HARQ}}\) can be regarded as a fixed latency \text{C}~\cite{skocaj2023data}. According to \cite{9964110}, the transmission process is modeled as an M/M/1 queue. The resource scheduling latency is viewed as the queuing time, while the data transmission latency is treated as the service time. It is assumed that data arrivals follow a Poisson distribution with parameter \(\lambda_1\), and the service time follows an exponential distribution with parameter \(\lambda_2\), denote \(\tau_{\text{tx}}=\tau_{\text{sch}} + \tau_{\text{data}} + \text{C}\) represents the processing and transmission latency, while the arrival of data packets can be represented as a Poisson process
\begin{multline} \label{eq: probabilities}
    P(N=n|\tau_{\text{tx}}=t) = \frac{P(N=n)\,P(\tau_{\text{tx}}|N=n)}{P(\tau_{\text{tx}}=t)} \\ = \frac{1}{P(\tau_{\text{tx}}=t)}\,P_n\,\frac{\lambda_2\,\mathrm{e}^{-\lambda_2\,t}\,(\lambda_2\,t)^{n-1}}{n!} \\ = \frac{(1-\frac{\lambda_1}{\lambda_2})\,\lambda_2\,\mathrm{e}^{-\lambda_2\,t}}{P(\tau_{\text{tx}}=t)}\frac{(\lambda_1\,t)^n}{n!}\,.
\end{multline}

According to the theory of Poisson processes in stochastic process theory, the inter-arrival time between Poisson-distributed events follows an exponential distribution. Therefore, the probability of the transmission latency can be expressed as
\begin{equation} \label{eq: transmission probability}
    P(\tau_{\text{tx}}) = (\lambda_2-\lambda_1)\,\mathrm{e}^{-(\lambda_2-\lambda_1)\,t}\,.
\end{equation}
According to~\autoref{eq: transmission probability}, the distribution of the transmission latency follows an exponential distribution with parameter \(\mu_1=\lambda_2-\lambda_1\).
For the retransmission process, assuming the retransmission success probability is \(p\), the probability of the first~\(n-1\) times transmissions failing and the~$n^{\text{th}}$ transmission succeeding can be expressed as \(P(N=n)=(1-p)^{n-1}p\). The number of retransmissions follow geometric distribution~\(N\sim \text{GE}(p)\), where \(p\) depends on the block error rate (BLER). Hence, the retransmission time~$\tau_{\text{rtx}}$ follows the exponential distribution with rate parameter~$\mu_{\text{2}}$.
Considering that the BLER is generally small, the focus is on cases where the number of retransmissions is 1. Thus,~\autoref{eq: RAN latency} can be simplified to
\begin{equation}
    l_{\text{radio}}=\tau_l \approx \tau_{\text{tx}} + \tau_{\text{rtx}_1} p_1\,.
\end{equation}
Based on the above derivation, the random variable \(l_\text{radio}\) can be viewed as the sum of two independent random variables, \(\tau_{\text{tx}}\) and \(\tau_{\text{rtx}_1}\) that follow exponential distributions. Therefore, its probability density function can be expressed as
\begin{multline} \label{eq: pdf of radio latency}
    f_{\text{radio}}(\tau_l) = \int_{-\infty}^{\infty} p_{\text{tx}}(\tau)\,p_{\text{rtx}_1}(\tau_l - \tau)\,\text{d}\tau 
    \\  = \int_{0}^{\tau_l} \mu_1\,\mathrm{e}^{-\mu_1\,\tau}\,\mu_2\,\mathrm{e}^{-\mu_2\,(\tau_l-\tau)} \,\text{d}\tau 
    \\ = \mu_1\,\mu_2\,\mathrm{e}^{-\mu_2\,\tau_l} \int_{0}^{\tau_l} \mathrm{e}^{-(\mu_1-\mu_2)\,\tau} \,\text{d}\tau\,. 
\end{multline}
Without loss of generality, it is assumed that the retransmission latency is approximately equal to the latency of the first transmission. Relations~\(\tau_{\text{tx}} \sim \mathrm{Exp}(\mu_1)\),~\(\tau_{\text{rtx}_1} \sim \mathrm{Exp}(\mu_2)\),~\(\mu = \mu_1 = \mu_2\) can be established.~\autoref{eq: pdf of radio latency} can be rewritten as
\begin{equation}
    f_{\text{radio}}(\tau_l) = \mu^2\,\mathrm{e}^{-\mu\,\tau_l} \int_{0}^{\tau_l} 1 \,\text{d}\tau = \mu^2\,\tau_l\, \mathrm{e}^{-\mu\,\tau_l}\,,
\end{equation}
which is equivalent to the probability distribution of \(l_{\text{radio}} \sim \mathrm{Gamma}(2, \mu)\). Therefore, the communication latency statistical model for 5G E2E theoretically follows a \textbf{Gamma distribution}.
}

\bibliographystyle{IEEEtran}
\bibliography{reference}

\vspace{-33pt}
\begin{IEEEbiographynophoto}{Yongqi Zhao}
received the bachelor’s degree from the China University of Petroleum (East China), Qingdao, China, in 2019, and the master’s degree from Technical University of Braunschweig, Braunschweig, Germany, in 2022. He is currently pursuing the Ph.D. degree with the Institute of Automotive Engineering, Graz University of Technology, Graz, Austria, with a research focus on virtual testing of automated driving systems. While pursuing his master’s degree, he gained practical experience through internships with Momenta, Stuttgart, Germany, and Volkswagen Group, Wolfsburg, Germany.
\end{IEEEbiographynophoto}
\vspace{-33pt}
\begin{IEEEbiographynophoto}{Xinrui Zhang}
received the B.S. and the M.S. degrees in computer science and technology from Chang’an University, Xi’an, China, in 2019 and 2022, respectively. He is currently working toward the Ph.D. degree in transportation engineering with the School of Automotive Studies, Tongji University, Shanghai, China. His research interests include cooperative control, test and evaluation of cloud-based intelligent connected vehicles.
\end{IEEEbiographynophoto}
\vspace{-33pt}
\begin{IEEEbiographynophoto}{Tomislav Mihalj}
received his degree in Mechanical Engineering from the University of Zagreb, Croatia, in 2014, and earned his PhD from Graz University of Technology, Austria, in 2024. From 2014 to 2019, he worked as a Research Engineer at Virtual Vehicle, Graz, Austria, where he focused on the mechanical efficiency of combustion engines, vibration analysis, and crack propagation in wheel-rail contact. Between 2019 and 2024, he served as a University Project Assistant at the Institute of Automotive Engineering, Graz University of Technology, concentrating on the virtual verification of automated driving systems. Since 2024, he has been a Postdoctoral Researcher at the same institute, where he focuses on the verification and validation of driver assistance systems and supervises related research projects.
He has authored or co-authored 12 peer-reviewed publications on rail vehicles and driver assistance systems. He has also contributed to several national and EU-funded projects related to automated driving.
\end{IEEEbiographynophoto}
\vspace{-33pt}
\begin{IEEEbiographynophoto}{Martin Schabauer} is holding a bachelor’s and master’s degree in mechanical engineering from Graz University of Technology, Austria. Since 2024, he has been pursuing a PhD degree at the same institution. Since 2014, he has been a researcher and lecturer at the Institute of Automotive Engineering at Graz University of Technology. He is doing research in vehicle dynamics and tire mechanics, including modelling, simulation, testing, parameter identification, and validation. The main focus of his research is on vehicle multibody dynamics and semi-physical tire modeling. As a lecturer, his teaching also focuses on modeling, simulation and testing of vehicles and tires. 
\end{IEEEbiographynophoto}
\vspace{-33pt}
\begin{IEEEbiographynophoto}{Luis Putzer} received his bachelor's and master's degrees in Information and Computer Engineering from Graz University of Technology, Graz, Austria, in 2020 and 2023, respectively. During his master's studies at the Institute of Automotive Engineering, Graz University of Technology, his research focused on human-machine 
interaction, with an focus on virtual development and validation in Automated Driving Systems (ADS). He is currently working at the Volkswagen Group.
\end{IEEEbiographynophoto}
\vspace{-33pt}
\begin{IEEEbiographynophoto}{Erik Reichmann-Blaga} received the master’s degree in mechanical engineering from Graz University of Technology, Graz, Austria, in 2024. He gained practical experience through an internship in the e-Drive System Development Division at Mercedes-Benz G GmbH from October 2023 to May 2024. His technical interests include automotive systems, control software, and development tools such as Vector CANape and JIRA.
\end{IEEEbiographynophoto}
\vspace{-33pt}
\begin{IEEEbiographynophoto}{Ádám Boronyák} earned his bachelor's degree in Electrical Engineering from University of Pannonia, Veszprém in 2018 and master's degree in Electrical Engineering from the Budapest University of Technology and Economics (BME), Budapest, Hungary in 2020. From August 2020, he works at the Department of Automotive Technologies in BME as a department engineer. His main tasks are to study communication devices and their interfaces, to develop software for communication on the internet and to manage the entire process of software development.
\end{IEEEbiographynophoto}
\vspace{-33pt}
\begin{IEEEbiographynophoto}{András Rövid} graduated in computer engineering, Faculty of Electrical Engineering and Informatics, Technical University of Kosice, Kosice, Slovakia, in 2001. He received the Ph.D. degree in transportation sciences from the Budapest University of Technology and Economics (BUTE), Budapest, Hungary, in 2005. He is currently a senior research fellow and from 2019 the leader of the Perception Group at the Department of Automotive Technologies of BUTE. He has been author or co-author of over 100 publications. His main interest include image processing, 3D machine vision, sensor fusion.
\end{IEEEbiographynophoto}
\vspace{-33pt}
\begin{IEEEbiographynophoto}{Gábor Soós} earned his degree in Electrical Engineering from the Budapest University of Technology and Economics in 2008. In 2007, he joined T-Systems Hungary, where he worked on the modeling and implementation of reliable, advanced radio communication technologies. After obtaining his degree, he was responsible for the DECT radio planning of the newly built Mercedes-Benz factory in Kecskemét and served as the radio network planning engineer for the AUDI plant in Győr. He obtained his PhD in Informatics from the Budapest University of Technology and Economics in 2022.

Currently, he works on the operation and development of Magyar Telekom's mobile Core network and focuses on the integration of advanced Industry 4.0 and 5G technologies. He is a guest editor of the "Vehicle to Everything" special issue in the MDPI Electronics journal. His research interests include mobile network development, testing of high-reliability private cellular core systems, and the exploration of V2X technologies. He remains an active researcher, having contributed to approximately 32 publications over the past six years, which have been cited more than 800 times and reached a readership of 60,000.
\end{IEEEbiographynophoto}
\vspace{-33pt}
\begin{IEEEbiographynophoto}{Peizhi Zhang}
received the M.S. degree in vehicle engineering from Jilin University, Changchun, China, in 2014 and the Ph.D. degree in vehicle engineering from Tongji University, Shanghai, China, in 2023. From 2014 to 2017, he was a research assistant at the National Intelligent New Energy Vehicle Collaborative Innovation Center. He is currently researching multi-vehicle cooperative motion planning and control.
\end{IEEEbiographynophoto}
\vspace{-33pt}
\begin{IEEEbiographynophoto}{Lu Xiong}
received the B.E., M.E., and the Ph.D. degrees in vehicle engineering from the School of Automotive Studies, Tongji University, Shanghai, China, in 1999, 2002, and 2005, respectively. From November 2008 to 2009, he was a Postdoctoral Fellow with the Institute of Automobile Engineering and Vehicle Engines, University of Stuttgart, Germany, with Dr. Jochen Wiedemann. He is currently a Professor with Tongji University. He is also an Executive Director of the Institute of Intelligent and an Associate Director of the Clean Energy Automotive Engineering Center, Tongji University. His research interests include perception, decision and planning, dynamics control and state estimation and testing and evaluation of intelligent connected vehicles.
\end{IEEEbiographynophoto}
\vspace{-33pt}
\begin{IEEEbiographynophoto}{Jia Hu} (Senior Member, IEEE) is currently the Zhongte Distinguished Chair of cooperative automation with the College of Transportation Engineering, Tongji University. Before joining Tongji University, he was a Research Associate with the Federal Highway Administration (FHWA), USA. He is a member of the TRB (Division of the National Academies) Vehicle Highway Automation Committee, the Freeway Operation Committee, and the Simulation Subcommittee of Traffic Signal Systems Committee, and a member of the CAV Impact Committee and the Artificial Intelligence Committee of the ASCE Transportation and Development Institute. He is an Advisory Editorial Board Member of~\textit{Transportation Research Part C: Emerging Technologies}. He has been an Associate Editor of the IEEE Intelligent Vehicles Symposium since 2018 and the IEEE Intelligent Transportation Systems Conference since 2019. He is an Associate Editor of~\textit{Journal of Transportation Engineering} (American Society of Civil Engineers) and~\textit{Journal of Intelligent Transportation Systems}.
\end{IEEEbiographynophoto}
\vspace{-33pt}
\begin{IEEEbiographynophoto}{Arno Eichberger}
(Member, IEEE) received the degree in mechanical engineering and the Ph.D. degree (Hons.) in technical sciences from the Graz University of Technology, Graz, Austria, in 1995 and 1998, respectively. 

From 1998 to 2007, he was a employed with Magna Steyr Fahrzeugtechnik AG\&Company, Graz, where he dealt with different aspects of active and passive safety. Since 2007, he has been working with the Institute of Automotive Engineering, Graz University of Technology, dealing with driver assistance systems, vehicle dynamics, and suspensions. Since 2012, he has been an Associate Professor holding a “venia docendi” of automotive engineering.
\end{IEEEbiographynophoto}

\vfill

\end{document}